\pdfoutput=1
\documentclass[letterpaper]{article}
\usepackage{flairs}
\usepackage{times}
\usepackage{helvet}
\usepackage{courier}

\usepackage{amsmath,amssymb,amsthm}
\usepackage{thmtools, thm-restate}
\usepackage{algorithm}
\usepackage[noend]{algpseudocode}
\algrenewcommand{\algorithmicindent}{1.15em}
\algnewcommand{\algorithmicbreak}{\textbf{break}}
\algnewcommand{\algorithmicforeach}{\textbf{for each}}
\algnewcommand\Break{\algorithmicbreak}
\algdef{SE}[FOR]{ForEach}{EndForEach}[1]
	{\algorithmicforeach\ #1\ \algorithmicdo}
	{\algorithmicend\ \algorithmicforeach}
\algdef{SE}[DOWHILE]{DoWhile}{EndDoWhile}{\algorithmicdo}[1]{\algorithmicwhile\ #1}
\makeatletter
	\ifthenelse{\equal{\ALG@noend}{t}}{\algtext*{EndForEach}}{}
	\ifthenelse{\equal{\ALG@noend}{t}}{\algtext*{EndDoWhile}}{}
\makeatother
\usepackage{graphicx}
\usepackage{textcomp}
\usepackage{xcolor}
\usepackage{acronym}
\usepackage[capitalise]{cleveref}
\usepackage{csquotes}
\usepackage{tikz}
\usetikzlibrary{arrows,arrows.meta,automata,backgrounds,calc,patterns,positioning,shapes,shadows}

\newtheorem{theorem}{Theorem}
\newtheorem{proposition}{Proposition}

\newtheorem{definition}{Definition}
\newtheorem{example}{Example}

\crefname{algorithm}{Alg.}{Algs.}
\Crefname{algorithm}{Algorithm}{Algorithms}
\crefname{corollary}{Cor.}{Cors.}
\Crefname{corollary}{Corollary}{Corollaries}
\crefname{definition}{Def.}{Defs.}
\Crefname{definition}{Definition}{Definitions}
\crefname{example}{Ex.}{Exs.}
\Crefname{example}{Example}{Examples}
\crefname{proposition}{Prop.}{Props.}
\Crefname{proposition}{Proposition}{Propositions}
\crefname{section}{Sec.}{Secs.}
\Crefname{section}{Section}{Sections}
\crefname{theorem}{Thm.}{Thms.}
\Crefname{theorem}{Theorem}{Theorems}

\acrodef{bn}[BN]{Bayesian network}
\acrodef{cp}[CP]{colour passing}
\acrodef{cpr}[ACP]{advanced colour passing}
\acrodef{crv}[CRV]{counting randvar}
\acrodef{deft}[DEFT]{detection of exchangeable factors}
\acrodef{fg}[FG]{factor graph}
\acrodef{ljt}[LJT]{lifted junction tree}
\acrodef{lv}[logvar]{logical variable}
\acrodef{lve}[LVE]{lifted variable elimination}
\acrodef{mn}[MN]{Markov network}
\acrodef{pcrv}[PCRV]{parameterised CRV}
\acrodef{pf}[parfactor]{parametric factor}
\acrodef{pfg}[PFG]{parametric factor graph}
\acrodef{prv}[PRV]{parameterised randvar}
\acrodef{rv}[randvar]{random variable}
\acrodef{ve}[VE]{variable elimination}
\acrodef{wl}[WL]{Weisfeiler-Leman}

\crefname{algorithm}{Alg.}{Algs.}
\Crefname{algorithm}{Algorithm}{Algorithms}
\crefname{definition}{Def.}{Defs.}

\newcommand{\alginput}[1]{\hspace*{\algorithmicindent} \textbf{Input:} #1}
\newcommand{\algoutput}[1]{\hspace*{\algorithmicindent} \textbf{Output:} #1}
\newcommand{\abs}[1]{\lvert #1 \rvert}
\newcommand{\cnt}{\ensuremath{\mathrm{count}}}

\newcommand{\unique}{\ensuremath{\mathrm{unique}}}

\definecolor{myyellow}{RGB}{247,192,26}
\definecolor{myblue}{RGB}{37,122,164}
\definecolor{mygreen}{RGB}{78,155,133}
\definecolor{mypurple}{RGB}{86,51,94}

\definecolor{newblue}{RGB}{50,113,173}
\definecolor{newred}{RGB}{222,32,36}
\definecolor{newgreen}{RGB}{70,165,69}
\definecolor{newpurple}{RGB}{140,69,152}

\definecolor{cborange}{RGB}{230,159,0}
\definecolor{cbblue}{RGB}{30,136,229}
\definecolor{cbbluedark}{RGB}{46,37,133}
\definecolor{cbpurple}{RGB}{170,68,153}
\definecolor{cbgreen}{RGB}{0,77,64}
\definecolor{cbgreenlight}{RGB}{93,168,153}
\definecolor{cbbrown}{RGB}{126,41,84}

\pgfdeclarelayer{bg}
\pgfsetlayers{bg,main}

\tikzset{
	rv/.style={draw, ellipse},
	pf/.style={draw, rectangle, fill = gray!30},
	arc/.style = {->, >={[round,sep]Stealth}},
}

\newcommand\factor[6]{
	\node[pf, #1=#3 of #2, label={#4:{#5}}](#6) {};
}

\begin{document}

\title{\vspace*{1cm} Efficient Detection of Exchangeable Factors in Factor Graphs\thanks{Extended
version of paper accepted to the Proceedings of the 37th International FLAIRS Conference (FLAIRS-24).}}

\author{
	Malte Luttermann\textsuperscript{1},
	Johann Machemer\textsuperscript{2} \and
	Marcel Gehrke\textsuperscript{2} \\
	\textsuperscript{1} German Research Center for Artificial Intelligence (DFKI), Lübeck, Germany \\
	\textsuperscript{2} Institute of Information Systems, University of Lübeck, Lübeck, Germany \\
	malte.luttermann@dfki.de, johann.machemer@student.uni-luebeck.de, gehrke@ifis.uni-luebeck.de
}

\maketitle

\begin{abstract}
\begin{quote}
	To allow for tractable probabilistic inference with respect to domain sizes, lifted probabilistic inference exploits symmetries in probabilistic graphical models.
	However, checking whether two factors encode equivalent semantics and hence are exchangeable is computationally expensive.
	In this paper, we efficiently solve the problem of detecting exchangeable factors in a \acl{fg}.
	In particular, we introduce the \emph{\acf{deft}} algorithm, which allows us to drastically reduce the computational effort for checking whether two factors are exchangeable in practice.
	While previous approaches iterate all $O(n!)$ permutations of a factor's argument list in the worst case (where $n$ is the number of arguments of the factor), we prove that \ac{deft} efficiently identifies restrictions to drastically reduce the number of permutations and validate the efficiency of \ac{deft} in our empirical evaluation.
\end{quote}
\end{abstract}

\acresetall

\section{Introduction}
	Probabilistic graphical models compactly encode a full joint probability distribution as a factorisation and provide a well-founded formalism to reason under uncertainty, e.g., when performing automated planning and acting.
	Reasoning under uncertainty, however, might become computationally expensive when using propositional probabilistic models such as \aclp{bn}, \aclp{mn}, or \acp{fg}.
	In general, probabilistic inference (i.e., the computation of marginal distributions of \acp{rv} given observations for other \acp{rv}) scales exponentially with the number of \acp{rv} in the \acl{bn}, \acl{mn}, or \ac{fg}, respectively, in the worst case.
	To allow for tractable probabilistic inference (e.g., probabilistic inference requiring polynomial time) with respect to domain sizes of \aclp{lv}, lifted probabilistic inference algorithms exploit symmetries in a probabilistic graphical model by using a representative of indistinguishable individuals for computations.
	However, to exploit symmetries in a probabilistic graphical model, these symmetries must be detected first and therefore, we investigate the problem of efficiently detecting exchangeable factors (i.e., factors that encode the same underlying function regardless of the order of their arguments) in \acp{fg} in this paper.

	In previous work, \citeauthor{Poole2003a}~(\citeyear{Poole2003a}) introduces \acp{pfg} and \acl{lve} as a lifted inference algorithm operating on \acp{pfg}.
	Since then, \acl{lve} has been refined by many researchers~\cite{DeSalvoBraz2005a,DeSalvoBraz2006a,Milch2008a,Kisynski2009a,Taghipour2013a,Braun2018a}.
	To perform lifted probabilistic inference, the lifted representation (e.g., the \ac{pfg}) has to be obtained first.
	The commonly used \ac{cp} algorithm (initially named as \enquote{CompressFactorGraph}) can be used to obtain an equivalent lifted representation of an \ac{fg}~\cite{Kersting2009a,Ahmadi2013a}.
	Among other refinements~\cite{Luttermann2023b}, the \ac{cp} algorithm has been extended to \emph{\ac{cpr}}~\cite{Luttermann2024a}, which transforms a given \ac{fg} into a \ac{pfg} entailing equivalent semantics as the initial \ac{fg} and, in contrast to \ac{cp}, does not require exchangeable factors to have their potential mappings specified in a specific order.
	However, the offline-step required by \ac{cpr} to detect exchangeable factors is computationally expensive.

	To efficiently detect exchangeable factors in \acp{fg}, we contribute the \emph{\acf{deft}} algorithm and show in an in-depth theoretical analysis that \ac{deft} avoids the expensive computation of permutations in many practical settings, making the search for exchangeable factors feasible in practice.
	More specifically, we analyse both the complexity of previous approaches and the problem itself and then apply the theoretical insights to obtain \ac{deft} as a practical algorithm.
	To tackle the problem of detecting exchangeable factors, we make use of so-called buckets that count the occurrences of specific range values in an assignment for a (sub)set of a factor's arguments.
	We show that using buckets, the exchangeability of factors can be detected in a highly efficient manner, thereby allowing us to drastically reduce the number of permutations to be checked.

	The remaining part of this paper is structured as follows.
	We begin by introducing necessary background information and notations.
	Thereafter, we delve into the problem of detecting exchangeable factors in \acp{fg} and provide an in-depth theoretical analysis of the problem.
	Finally, we present the \ac{deft} algorithm to drastically reduce the required computational effort to detect exchangeable factors in practice and confirm its efficiency in an empirical evaluation.

\section{Background}
	We begin by defining \acp{fg} as undirected propositional probabilistic models and then continue to introduce the notion of exchangeable factors within \acp{fg}.
	An \ac{fg} compactly encodes a full joint probability distribution between \acp{rv} by decomposing it into a product of factors~\cite{Frey1997a,Kschischang2001a}.

	\begin{definition}[Factor Graph]
		An \emph{\ac{fg}} $G = (\boldsymbol V, \boldsymbol E)$ is an undirected bipartite graph with node set $\boldsymbol V = \boldsymbol R \cup \boldsymbol \Phi$ where $\boldsymbol R = \{R_1, \ldots, R_n\}$ is a set of variable nodes (\acp{rv}) and $\boldsymbol \Phi = \{\phi_1, \ldots, \phi_m\}$ is a set of factor nodes (functions).
		The term $\mathcal R(R_i)$ denotes the possible values (range) of a \ac{rv} $R_i$.
		There is an edge between a variable node $R_i$ and a factor node $\phi_j$ in $\boldsymbol E \subseteq \boldsymbol R \times \boldsymbol \Phi$ if $R_i$ appears in the argument list of $\phi_j$.
		A factor is a function that maps its arguments to a positive real number, called potential.
		The semantics of $G$ is given by
		$$
			P_G = \frac{1}{Z} \prod_{j=1}^m \phi_j(\mathcal A_j)
		$$
		with $Z$ being the normalisation constant and $\mathcal A_j$ denoting the \acp{rv} connected to $\phi_j$ (i.e., the arguments of $\phi_j$).
	\end{definition}

	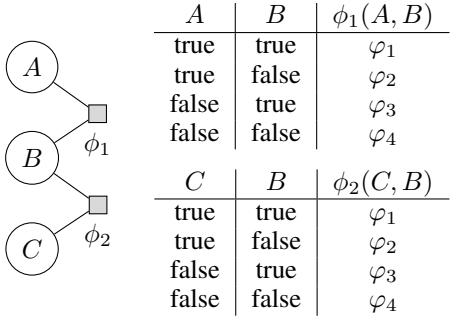
\begin{figure}[t]
		\centering
		\begin{tikzpicture}
			\node[circle, draw] (A) {$A$};
			\node[circle, draw] (B) [below = 0.5cm of A] {$B$};
			\node[circle, draw] (C) [below = 0.5cm of B] {$C$};
			\factor{below right}{A}{0.25cm and 0.5cm}{270}{$\phi_1$}{f1}
			\factor{below right}{B}{0.25cm and 0.5cm}{270}{$\phi_2$}{f2}
		
			\node[right = 0.5cm of f1, yshift=5.0mm] (tab_f1) {
				\begin{tabular}{c|c|c}
					$A$   & $B$   & $\phi_1(A,B)$ \\ \hline
					true  & true  & $\varphi_1$ \\
					true  & false & $\varphi_2$ \\
					false & true  & $\varphi_3$ \\
					false & false & $\varphi_4$ \\
				\end{tabular}
			};
		
			\node[right = 0.5cm of f2, yshift=-5.0mm] (tab_f2) {
				\begin{tabular}{c|c|c}
					$C$   & $B$   & $\phi_2(C,B)$ \\ \hline
					true  & true  & $\varphi_1$ \\
					true  & false & $\varphi_2$ \\
					false & true  & $\varphi_3$ \\
					false & false & $\varphi_4$ \\
				\end{tabular}
			};
		
			\draw (A) -- (f1);
			\draw (B) -- (f1);
			\draw (B) -- (f2);
			\draw (C) -- (f2);
		\end{tikzpicture}
		\caption{A toy example for an \ac{fg} consisting of three Boolean \acp{rv} $A$, $B$, and $C$ as well as two factors $\phi_1$ and $\phi_2$. The mappings of $\phi_1$ and $\phi_2$ are given in the respective tables on the right with $\varphi_1$, \dots, $\varphi_4 \in \mathbb{R}^+$.}
		\label{fig:example_fg}
	\end{figure}

	\begin{example} \label{ex:fg}
		\Cref{fig:example_fg} shows a toy example for an \ac{fg} consisting of three Boolean \acp{rv} $A$, $B$, and $C$ as well as two factors $\phi_1$ and $\phi_2$.
		Note that in this example, $\phi_1$ and $\phi_2$ encode identical potentials (depicted in the tables on the right) and hence are exchangeable.
		More specifically, it holds that $\phi_1(A,B) = \phi_2(C,B)$ for all inputs where $A$ and $C$ are assigned the same value.
		Consequently, we can say that $A$ and $C$ are exchangeable, enabling us to treat them equally during probabilistic inference.
	\end{example}

	In general, grouping together identically behaving objects and using only a single representative for each group is the core idea behind lifted inference~\cite{Niepert2014a}.
	Lifted inference algorithms exploit symmetries in a probabilistic graphical model by operating on \acp{pfg} which consist of \aclp{prv} and \aclp{pf}, representing sets of \acp{rv} and factors, respectively~\cite{Poole2003a}.
	Coming back to \cref{ex:fg}, recall that the semantics of the \ac{fg} $G$ depicted in \cref{fig:example_fg} is given by $P_G = \frac{1}{Z} \cdot \phi_1(A,B) \cdot \phi_2(C,B)$.
	As $A$ and $C$ (and hence $\phi_1$ and $\phi_2$) are exchangeable, we can use a single representative factor that represents a group of identically behaving factors (here $\phi_1$ and $\phi_2$) and take it to the power of the group size instead of multiplying each factor separately.

	Symmetries in \acp{fg} occur not only in our toy example but are highly relevant in various real world domains such as an epidemic domain where each individual person influences the probability of an epidemic in the same way---because the probability of having an epidemic depends on the number of sick people and not on individual people being sick.
	More specifically, the probability for an epidemic is the same if there are three sick people and the remaining people in the universe are not sick, independent of whether $alice$, $bob$, and $eve$ or $charlie$, $dave$, and $fred$ are sick, for example.
	Analogously, in a movie domain the popularity of an actor influences the success of a movie in the same way for each actor, in a research domain the quality of every publication influences the quality of a conference equally, and so on.

	To detect symmetries in an \ac{fg} and exploit them to speed up probabilistic inference, the \ac{cpr} algorithm~\cite{Luttermann2024a}, which generalises the \ac{cp} algorithm~\cite{Kersting2009a,Ahmadi2013a}, is the state of the art.
	\Ac{cpr} deploys a subroutine that checks whether two factors are exchangeable to find out which factors should be grouped together.
	Before we investigate how this subroutine can efficiently be realised, we give a formal definition of exchangeable factors.

	\begin{definition}[Exchangeable Factors] \label{def:exchangeable}
		Let $\phi_1(R_1, \ldots, R_n)$ and $\phi_2(R'_1, \ldots, R'_n)$ denote two factors in an \ac{fg} $G$.
		Then, $\phi_1$ and $\phi_2$ represent equivalent potentials if and only if there exists a permutation $\pi$ of $\{1, \ldots, n\}$ such that for all $r_1, \ldots, r_n \in \times_{i=1}^n \mathcal R(R_i)$ it holds that $\phi_1(r_1, \ldots, r_n) = \phi_2(r_{\pi(1)}, \ldots, r_{\pi(n)})$.
		Factors that represent equivalent potentials are called \emph{exchangeable}.
	\end{definition}

	Note that two factors must have the same number of arguments $n$ to be able to be exchangeable because otherwise they cannot encode the same underlying function.
	Further, it must hold that there exists a bijection $\tau \colon \{R_1, \ldots, R_n\} \to \{R'_1, \ldots, R'_n\}$ which maps each $R_i$ to an $R'_j$ such that $\mathcal R(R_i) = \mathcal R(R'_j)$.
	In other words, the ranges of the arguments of two exchangeable factors must be the same to ensure that the two functions encoded by the factors are defined over the same function domain.

	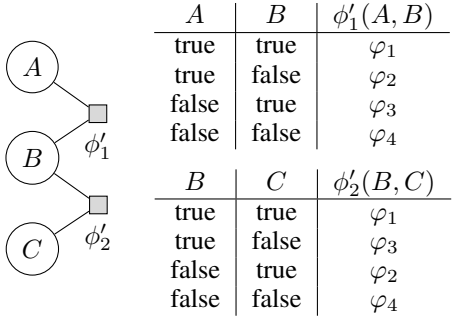
\begin{figure}[t]
		\centering
		\begin{tikzpicture}
			\node[circle, draw] (A) {$A$};
			\node[circle, draw] (B) [below = 0.5cm of A] {$B$};
			\node[circle, draw] (C) [below = 0.5cm of B] {$C$};
			\factor{below right}{A}{0.25cm and 0.5cm}{270}{$\phi'_1$}{f1}
			\factor{below right}{B}{0.25cm and 0.5cm}{270}{$\phi'_2$}{f2}
		
			\node[right = 0.5cm of f1, yshift=5.0mm] (tab_f1) {
				\begin{tabular}{c|c|c}
					$A$   & $B$   & $\phi'_1(A,B)$ \\ \hline
					true  & true  & $\varphi_1$ \\
					true  & false & $\varphi_2$ \\
					false & true  & $\varphi_3$ \\
					false & false & $\varphi_4$ \\
				\end{tabular}
			};
		
			\node[right = 0.5cm of f2, yshift=-5.0mm] (tab_f2) {
				\begin{tabular}{c|c|c}
					$B$   & $C$   & $\phi'_2(B,C)$ \\ \hline
					true  & true  & $\varphi_1$ \\
					true  & false & $\varphi_3$ \\
					false & true  & $\varphi_2$ \\
					false & false & $\varphi_4$ \\
				\end{tabular}
			};
		
			\draw (A) -- (f1);
			\draw (B) -- (f1);
			\draw (B) -- (f2);
			\draw (C) -- (f2);
		\end{tikzpicture}
		\caption{Another toy example for an \ac{fg} consisting of three Boolean \acp{rv} $A$, $B$, and $C$ as well as two factors $\phi_1$ and $\phi_2$. Observe that the full joint probability distribution encoded by the illustrated \ac{fg} is the same as the probability distribution encoded by the \ac{fg} depicted in \cref{fig:example_fg}.}
		\label{fig:example_fg_permuted}
	\end{figure}

	\begin{example} \label{ex:exchangeable}
		Take a look at the \ac{fg} $G$ illustrated in \cref{fig:example_fg_permuted}.
		$G$ entails equivalent semantics as the \ac{fg} depicted in \cref{fig:example_fg} because $\phi_1(A,B) = \phi'_1(A,B)$ and $\phi_2(C,B) = \phi'_2(B,C)$ for all possible assignments of $A$, $B$, and $C$.
		More specifically, in $\phi'_2$, the order of the arguments $B$ and $C$ is swapped compared to their order in $\phi_2$.
		The potentials, however, are still the same as $\phi_2(C = \mathrm{true}, B = \mathrm{false}) = \phi'_2(B = \mathrm{false}, C = \mathrm{true}) = \varphi_2$ and $\phi_2(C = \mathrm{false}, B = \mathrm{true}) = \phi'_2(B = \mathrm{true}, C = \mathrm{false}) = \varphi_3$.
		Consequently, all of the four factors $\phi_1$, $\phi_2$, $\phi'_1$, and $\phi'_2$ are exchangeable.
	\end{example}

	As we have seen in \cref{ex:exchangeable}, it is not necessarily the case that the tables of the input-output mappings of two exchangeable factors are identical.
	For practical applications, it is far too strong an assumption to make that the tables of exchangeable factors are always ordered such that they read identical values from top to bottom.
	Therefore, we next briefly recap the previous approach to detect exchangeable factors and afterwards continue to provide an in-depth investigation of the problem of detecting exchangeable factors in \acp{fg}, allowing us to pour the gained insights into the \ac{deft} algorithm to efficiently detect exchangeable factors in practice.

\section{Efficient Detection of Exchangeable Factors Using Buckets}
	A straightforward approach to check whether two factors $\phi_1(R_1, \ldots, R_n)$ and $\phi_2(R'_1, \ldots, R'_n)$ are exchangeable is to loop over all permutations of one of the factors' argument lists, say the argument list of $\phi_2$, and then check for each permutation whether $\phi_1$ and $\phi_2$ map to the same potential for all assignments of their arguments (i.e., whether both tables read identical values from top to bottom after the rearrangement of $\phi_2$'s arguments according to the permutation).
	\citeauthor{Luttermann2024a}~(\citeyear{Luttermann2024a}) give a more efficient approach where histograms (which we call \emph{buckets} in this paper) entailed by so-called \aclp{crv} are used to enforce a necessary condition for two factors to be exchangeable, thereby allowing to heavily prune the search space and check the permutations only after the initial filtering succeeds.
	Nevertheless, the approach still checks all $n!$ permutations after the initial filtering succeeds.
	To provide a theoretical analysis and then show how the approach can be drastically improved, we first introduce the notion of a bucket.

	\begin{definition}[Bucket]
		Let $\phi(R_1, \ldots, R_n)$ be a factor and let $\boldsymbol S \subseteq \{R_1, \ldots, R_n\}$ denote a subset of $\phi$'s arguments such that $\mathcal R(R_i) = \mathcal R(R_j)$ for all $R_i, R_j \in \boldsymbol S$.
		Further, let $\mathcal V$ denote the range of the elements in $\boldsymbol S$ (identical for all $R_i \in \boldsymbol S$).
		Then, a \emph{bucket} $b$ entailed by $\boldsymbol S$ is a set of tuples $\{(v_i, n_i)\}_{i = 1}^{\abs{\mathcal V}}$, $v_i \in \mathcal V$, $n_i \in \mathbb{N}$, and $\sum_i n_i = \abs{\boldsymbol S}$, such that $n_i$ specifies the number of occurrences of value $v_i$ in an assignment for all \acp{rv} in $\boldsymbol S$.
		A shorthand notation for $\{(v_i, n_i)\}_{i = 1}^{\abs{\mathcal V}}$ is $[n_1, \dots, n_{\abs{\mathcal V}}]$.
		In abuse of notation, we denote by $\phi(b)$ the multiset of potentials the assignments represented by $b$ are mapped to by $\phi$.
	\end{definition}

	\begin{example}
		Consider the factor $\phi'_1(A,B)$ from \cref{fig:example_fg_permuted} and let $\boldsymbol S = \{A, B\}$ with $\mathcal R(A) = \mathcal R(B) = \{\mathrm{true}, \mathrm{false}\}$.
		Then, $\boldsymbol S$ entails the three buckets $\{(\mathrm{true}, 2), \allowbreak (\mathrm{false}, 0)\}$, $\{(\mathrm{true}, 1), \allowbreak (\mathrm{false}, 1)\}$, and $\{(\mathrm{true}, 0), \allowbreak (\mathrm{false}, 2)\}$---or $[2, 0]$, $[1, 1]$, and $[0, 2]$ in shorthand notation.
		According to the mappings depicted in the table shown in \cref{fig:example_fg_permuted}, it holds that $\phi'_1([2, 0]) = \langle \varphi_1 \rangle$, $\phi'_1([1, 1]) = \langle \varphi_2, \varphi_3 \rangle$, and $\phi'_1([0, 2]) = \langle \varphi_4 \rangle$.
	\end{example}

	In other words, buckets count the occurrences of specific range values in an assignment for a subset of a factor's arguments.
	Hence, every bucket may be mapped to multiple potentials while every potential (row in the table of mappings of a factor) belongs to exactly one bucket.

\subsection{Buckets over Multiple Ranges}
	The reason we define buckets over a subset of a factor's arguments (instead of over the whole argument list) is that we are looking for a permutation of arguments having the same range such that the potential mappings of two factors are identical.
	The following example illustrates this point.

	\begin{example}
		Consider $\phi_1(R_1, R_2, R_3)$ and $\phi_2(R_4, R_5, R_6)$ with $\mathcal R(R_1) = \mathcal R(R_2) = \mathcal R(R_4) = \mathcal R(R_5) = \{\mathrm{true}, \mathrm{false}\}$ and $\mathcal R(R_3) = \mathcal R(R_6) = \{\mathrm{low}, \mathrm{medium}, \mathrm{high}\}$.
		To be able to obtain identical tables of potential mappings for $\phi_1$ and $\phi_2$, $R_3$ and $R_6$ must be located at the same position in the respective argument list of $\phi_1$ and $\phi_2$ because they are the only arguments having a non-Boolean range.
		Therefore, we compare the buckets entailed by $\boldsymbol S_1 = \{R_1, R_2\}$ and $\boldsymbol S_2 = \{R_4, R_5\}$ separately from the buckets entailed by $\boldsymbol S_3 = \{R_3\}$ and $\boldsymbol S_4 = \{R_6\}$ when checking whether $\phi_1$ and $\phi_2$ are exchangeable.
	\end{example}

	As we are interested in using buckets to check whether arguments with the same range can be permuted such that the tables of two factors are identical, from now on we consider only factors with arguments having the same range.
	If this simplification does not hold, i.e., if a factor contains arguments with different ranges $\mathcal R_1, \dots, \mathcal R_k$, the buckets for all arguments with range $\mathcal R_i$ must be compared separately for each $i = 1, \dots, k$.
	Having introduced the notion of buckets, we are now able to investigate the complexity of detecting exchangeable factors with the help of buckets.

\subsection{Properties of Buckets}
	Before we take a closer look at using buckets for detecting exchangeable factors, we briefly state the complexity of the problem of detecting exchangeable factors.

	\begin{theorem}
		Let $\phi_1(R_1, \ldots, R_n)$ and $\phi_2(R'_1, \ldots, R'_n)$ denote two factors.
		The number of table comparisons needed to check whether $\phi_1$ and $\phi_2$ are exchangeable is in $\mathcal O(n!)$.
	\end{theorem}
	\begin{proof}
		According to \cref{def:exchangeable}, there must exist a permutation of $\{1, \ldots, n\}$ such that $\phi_1$ and $\phi_2$ map to the same potential for all assignments of their arguments.
		As there are $n!$ permutations of $\{1, \ldots, n\}$ in total and we have to check every single permutation in the worst case, checking whether $\phi_1$ and $\phi_2$ are exchangeable is in $\mathcal O(n!)$.
	\end{proof}

	Even though we have to compare the two tables $n!$ times in the worst case, we can still drastically reduce the computational effort in many practical settings.
	The current state of the art, as presented by \citeauthor{Luttermann2024a}~(\citeyear{Luttermann2024a}), uses buckets as a necessary pre-condition that must be fulfilled for two factors to be exchangeable.

	\begin{proposition}[\citeauthor{Luttermann2024a},~\citeyear{Luttermann2024a}]
		Let $\phi_1$ and $\phi_2$ denote two exchangeable factors.
		Then, $\phi_1$ and $\phi_2$ are defined over the same function domain and hence their arguments entail the same buckets.
	\end{proposition}

	\begin{proposition}[\citeauthor{Luttermann2024a},~\citeyear{Luttermann2024a}] \label{prop:bucket_comparison}
		Let $\phi_1$ and $\phi_2$ denote two factors.
		If there exists a bucket $b$ such that $\phi_1(b) \neq \phi_2(b)$, then $\phi_1$ and $\phi_2$ are not exchangeable.
	\end{proposition}

	The approach of checking whether $\phi_1$ and $\phi_2$ are exchangeable is to compute all buckets entailed by $\phi_1$ and $\phi_2$, respectively, and compare them pairwise to check whether they are mapped to identical multisets of potentials.
	If this check fails, $\phi_1$ and $\phi_2$ are not exchangeable and hence no permutations are computed, otherwise it is checked whether $\phi_1$ and $\phi_2$ have identical tables of potential mappings for all permutations of the argument list of either $\phi_1$ or $\phi_2$.

	The computation of the buckets has no impact on the worst-case complexity as the number of buckets is always smaller than the number of rows in the table of potential mappings, which must be compared anyway.
	As each potential value belongs to exactly one bucket, the effort for comparing all buckets to each other is identical to the effort of comparing the two tables to each other.
	Therefore, applying a pre-pruning strategy that uses buckets to check whether two factors might be exchangeable at all reduces the computational effort in many practical settings.
	We next show that the approach can be further improved to drastically speed up the detection of exchangeable factors in practice.

	A crucial observation is that two factors are exchangeable if and only if their buckets are identical under consideration of the order of the values in the buckets.
	In particular, we denote by $\phi^{\succ}(b)$ the ordered multiset of potentials a bucket $b$ is mapped to by $\phi$ (in order of their appearance in the table of $\phi$) and then prove this insight in the following theorem.

	\begin{theorem} \label{th:bucket_orders}
		Let $\phi_1$ and $\phi_2$ denote two factors.
		Then, $\phi_1$ and $\phi_2$ are exchangeable if and only if there exists a permutation of their arguments such that $\phi_1^{\succ}(b) = \phi_2^{\succ}(b)$ for all buckets $b$ entailed by the arguments of $\phi_1$ and $\phi_2$.
	\end{theorem}
	\begin{proof}
		Let $\phi_1$ and $\phi_2$ denote two factors.
		For the first direction, it holds that $\phi_1$ and $\phi_2$ are exchangeable.
		According to \cref{def:exchangeable}, there exists a permutation of $\phi_2$'s arguments such that $\phi_1$ and $\phi_2$ have identical tables of potential mappings.
		Then, we have $\phi_1^{\succ}(b) = \phi_2^{\succ}(b)$ for every bucket $b$ as both tables read identical values from top to bottom.

		For the second direction, it holds that $\phi_1^{\succ}(b) = \phi_2^{\succ}(b)$ for all buckets $b$ entailed by the arguments of $\phi_1$ and $\phi_2$.
		As the order of the values in the buckets is identical for $\phi_1$ and $\phi_2$, converting the buckets back to tables of potential mappings results in identical tables for $\phi_1$ and $\phi_2$ and consequently, $\phi_1$ and $\phi_2$ are exchangeable.
	\end{proof}

	\begin{figure*}[t]
		\centering
		\begin{tikzpicture}
			\node (tab_f1) {
				\begin{tabular}{c|c|c|c|c}
					$R_1$ & $R_2$ & $R_3$ & $\phi_1$ & $b$ \\ \hline
					true  & true  & true  & $\varphi_1$           & {\color{cborange}$[3,0]$} \\
					true  & true  & false & $\varphi_2$           & {\color{cbbluedark}$[2,1]$} \\
					true  & false & true  & $\varphi_3$           & {\color{cbbluedark}$[2,1]$} \\
					true  & false & false & $\varphi_4$           & {\color{cbpurple}$[1,2]$} \\
					false & true  & true  & $\varphi_5$           & {\color{cbbluedark}$[2,1]$} \\
					false & true  & false & $\varphi_6$           & {\color{cbpurple}$[1,2]$} \\
					false & false & true  & $\varphi_6$           & {\color{cbpurple}$[1,2]$} \\
					false & false & false & $\varphi_7$           & {\color{cbgreen}$[0,3]$} \\
				\end{tabular}
			};
		
			\node[right = 0.2cm of tab_f1] (buckets) {
				\begin{tabular}{c|c|c}
					$b$                         & $\phi_1^{\succ}(b)$                                       & $\phi_2^{\succ}(b)$ \\ \hline
					{\color{cborange}$[3,0]$}   & $\langle \varphi_1 \rangle$                       & $\langle \varphi_1 \rangle$ \\
					{\color{cbbluedark}$[2,1]$} & $\langle \varphi_2, \varphi_3, \varphi_5 \rangle$ & $\langle \varphi_3, \varphi_5, \varphi_2 \rangle$ \\
					{\color{cbpurple}$[1,2]$}   & $\langle \varphi_4, \varphi_6, \varphi_6 \rangle$ & $\langle \varphi_6, \varphi_4, \varphi_6 \rangle$ \\
					{\color{cbgreen}$[0,3]$}    & $\langle \varphi_7 \rangle$                       & $\langle \varphi_7 \rangle$ \\
				\end{tabular}
			};
		
			\node[right = 0.2cm of buckets] (tab_f2) {
				\begin{tabular}{c|c|c|c|c}
					$R_4$ & $R_5$ & $R_6$ & $\phi_2$ & $b$ \\ \hline
					true  & true  & true  & $\varphi_1$           & {\color{cborange}$[3,0]$} \\
					true  & true  & false & $\varphi_3$           & {\color{cbbluedark}$[2,1]$} \\
					true  & false & true  & $\varphi_5$           & {\color{cbbluedark}$[2,1]$} \\
					true  & false & false & $\varphi_6$           & {\color{cbpurple}$[1,2]$} \\
					false & true  & true  & $\varphi_2$           & {\color{cbbluedark}$[2,1]$} \\
					false & true  & false & $\varphi_4$           & {\color{cbpurple}$[1,2]$} \\
					false & false & true  & $\varphi_6$           & {\color{cbpurple}$[1,2]$} \\
					false & false & false & $\varphi_7$           & {\color{cbgreen}$[0,3]$} \\
				\end{tabular}
			};
		\end{tikzpicture}
		\caption{Two exchangeable factors $\phi_1(R_1, R_2, R_3)$ (abbreviated as $\phi_1$) and $\phi_2(R_4, R_5, R_6)$ (abbreviated as $\phi_2$) and their corresponding buckets. Rearranging, for example, the arguments of $\phi_2$ such that they appear in order $R_5$, $R_6$, $R_4$ results in identical tables of potential mappings for $\phi_1$ and $\phi_2$.}
		\label{fig:example_deg_freedom}
	\end{figure*}

	To make use of \cref{th:bucket_orders}, the basic idea is that keeping an order for the elements in each bucket allows us to detect which arguments must be swapped for two buckets to be able to exactly match.
	The next example illustrates this idea.

	\begin{example}
		Let us take a look at \cref{fig:example_deg_freedom}.
		When comparing the buckets for $\phi_1$ and $\phi_2$, we can for example observe that the bucket $[2,1]$ is mapped to the same multiset of values, i.e., $\phi_1([2,1]) = \phi_2([2,1])$, but the values are ordered differently if we order them according to their appearance in the tables, i.e., $\phi_1^{\succ}([2,1]) \neq \phi_2^{\succ}([2,1])$.
		To obtain identical orders of values for both $\phi_1^{\succ}([2,1])$ and $\phi_2^{\succ}([2,1])$, e.g., $\varphi_2$ has to be swapped to the first position in $\phi_2^{\succ}([2,1])$ (i.e., $\varphi_2$ and $\varphi_3$ have to be swapped).
		Consequently, we can swap the assignments of $\phi_2$ that map to $\varphi_2$ and $\varphi_3$ to locate $\varphi_2$ at the first position in the bucket of $\phi_2$.
		In particular, we know that $\varphi_2$ belongs to the assignment $(\mathrm{false}, \mathrm{true}, \mathrm{true})$ and $\varphi_3$ belongs to the assignment $(\mathrm{true}, \mathrm{true}, \mathrm{false})$, i.e., $R_4$ and $R_6$ must be swapped in the argument list of $\phi_2$.
		This procedure of swapping two arguments can then be repeated until the buckets of $\phi_1$ and $\phi_2$ are identical (or until it is clear that the buckets cannot be made identical by swapping arguments).
	\end{example}

	In this example, the order of the potentials in the bucket $[2,1]$ uniquely determines which values and hence which arguments must be swapped to obtain identical buckets.
	As soon as the buckets contain duplicate values (such as it is the case for the bucket $[1,2]$ in \cref{fig:example_deg_freedom}), however, there might be multiple candidates for the next swap.
	We formalise this observation in the following definition.

	\begin{definition}[Degree of Freedom]
		Let $\phi(R_1, \ldots, R_n)$ be a factor and let $b$ be a bucket entailed by $\boldsymbol S \subseteq \{R_1, \ldots, R_n\}$.
		The \emph{degree of freedom} of $b$ is given by
		$$
			\mathcal F(b) = \prod_{\varphi \in \unique(\phi(b))} \cnt(\phi(b), \varphi)!
		$$
		where $\unique(\phi(b))$ denotes the set of unique potentials in $\phi(b)$ and $\cnt(\phi(b), \varphi)$ denotes the number of occurrences of potential $\varphi$ in $\phi(b)$.
	\end{definition}

	Then, the degree of freedom of a factor $\phi$ is defined as $\mathcal F(\phi) = \min_{b \in \{b \mid b \in \mathcal B(\phi) \land \abs{\phi(b)} > 1\}} \mathcal F(b)$, where $\mathcal B(\phi)$ denotes the set of all buckets entailed by $\phi$'s arguments.
	We consider only buckets that are mapped to at least two potential values because buckets which are mapped to a single potential correspond to all arguments being assigned the same value and hence, there is no need to swap arguments.
	Consequently, we define $\mathcal F(\phi) = 1$ if $\phi$ does not entail any buckets that are mapped to at least two potential values (which is only the case for factors having less than two arguments).

	\begin{example}
		Consider the factor $\phi_1$ given in \cref{fig:example_deg_freedom}.
		It holds that $\mathcal B(\phi_1) = \{[3,0], [2,1], [1,2], [0,3]\}$ and $\mathcal F([3,0]) = 1!$, $\mathcal F([2,1]) = 1! \cdot 1! \cdot 1!$, $\mathcal F([1,2]) = 1! \cdot 2!$, and $\mathcal F([0,3]) = 1!$.
		The buckets $[2,1]$ and $[1,2]$ are the only ones being mapped to a multiset consisting of more than one element and thus, it holds that $\mathcal F(\phi_1) = \min_{b \in \{[2,1], [1,2]\}} \mathcal F(b) = 1$.
	\end{example}

	Note that we take the minimum degree of freedom of the buckets as the degree of freedom of a factor.
	Let us again take a look at $\phi_1$ from \cref{fig:example_deg_freedom} to explain the reason for the minimum operation.
	Observe that even though the bucket $[1,2]$ contains a duplicate value and hence, we do not immediately know which of the two $\varphi_6$ values belongs to which position (and therefore, we do not uniquely know which arguments to swap), the bucket $[2,1]$ fixes the order of the arguments already.
	In other words, it is sufficient to swap the arguments according to the order induced by the bucket $[2,1]$ and afterwards check for all other buckets $b$ whether $\phi_1^{\succ}(b) = \phi_2^{\succ}(b)$.
	We next make use of this observation to present our main result and give a more precise complexity analysis of the problem of detecting exchangeable factors.
	\begin{theorem} \label{th:complexity}
		Let $\phi_1(R_1, \ldots, R_n)$ and $\phi_2(R'_1, \ldots, R'_n)$ denote two factors and let $d = \min\{\mathcal F(\phi_1), \mathcal F(\phi_2)\}$.
		The number of table comparisons needed to check whether $\phi_1$ and $\phi_2$ are exchangeable is in $\mathcal O(d)$.
	\end{theorem}
	\begin{proof}
		From \cref{prop:bucket_comparison}, we know that if $\mathcal F(\phi_1) \neq \mathcal F(\phi_2)$, $\phi_1$ and $\phi_2$ are not exchangeable and hence we do not need to try any permutations of arguments for table comparisons.
		Thus, we can assume that $\mathcal F(\phi_1) = \mathcal F(\phi_2)$ for the remaining part of this proof.
		Let $d = \mathcal F(\phi_1) = \mathcal F(\phi_2)$.
		Then, it holds that there exists a bucket $b'$ with $\mathcal F(b') = d$.
		If $\phi_1(b') \neq \phi_2(b')$, $\phi_1$ and $\phi_2$ are not exchangeable and we are done, so let $\phi_1(b') = \phi_2(b')$.
		From \cref{th:bucket_orders}, we know that for $\phi_1$ and $\phi_2$ to be able to be exchangeable, there must exist a permutation of their arguments such that $\phi_1^{\succ}(b) = \phi_2^{\succ}(b)$ for all buckets $b$, including $b'$.
		Hence, it is sufficient to try all permutations of arguments of, say $\phi_2$, for table comparison that are possible according to bucket $b'$ to find out whether $\phi_1$ and $\phi_2$ are exchangeable.
		Note that $\phi_1^{\succ}(b') = \langle \varphi_1, \dots, \varphi_{\ell} \rangle$ restricts the possible permutations of $\phi_2$'s arguments as the potential $\varphi_1$ must be placed at the first position in $\phi_2^{\succ}(b')$, and analogously for the other potentials.
		Therefore, each potential $\varphi_i$ can be placed at $\cnt(\phi_2(b'), \varphi_i)$ different positions in $\phi_2^{\succ}(b')$ and we have to try all permutations of these positions, resulting in $\cnt(\phi_2(b'), \varphi_i)!$ permutations that must be checked for each unique potential $\varphi_i$.
		Consequently, the number of permutations permitted by $b'$ is given by $d$, which completes the proof.
	\end{proof}

	Intuitively, \cref{th:complexity} tells us that the number of different potential values within the buckets of a factor determines the amount of permutations that must be iterated over in the worst case.
	Fortunately, for any factor $\phi(R_1, \dots, R_n)$, its potential values are mostly not identical in practice.
	Thus, $\mathcal F(\phi)$ is significantly smaller than $n!$ in most practical settings.
	In particular, $\mathcal F(\phi)$ is upper-bounded by $n!$.

	\begin{restatable}{corollary}{upperboundCorollary}
		Let $\phi(R_1, \ldots, R_n)$.
		It holds that $\mathcal F(\phi) \leq n!$.
	\end{restatable}

	Next, we exploit the theoretical insights from this section to efficiently detect exchangeable factors in practice.

\section{The \acs{deft} Algorithm}
	We now present the \ac{deft} algorithm to efficiently detect exchangeable factors in practical applications.
	\Cref{alg:deft} presents the entire \ac{deft} algorithm, which proceeds as follows to check whether two given factors $\phi_1(R_1, \dots, R_n)$ and $\phi_2(R'_1, \dots, R'_m)$ are exchangeable.
	\begin{algorithm}[t]
		\caption{Detection of Exchangeable Factors (DEFT)}
		\label{alg:deft}
		\alginput{Factors $\phi_1(R_1, \dots, R_n)$ and $\phi_2(R'_1, \dots, R'_m)$.} \\
		\algoutput{$\mathrm{true}$ if $\phi_1$ and $\phi_2$ are exchangeable, else $\mathrm{false}$.}
		\begin{algorithmic}[1]
			\If{$n \neq m \lor \mathcal B(\phi_1) \neq \mathcal B(\phi_2)$}
				\State \Return $\mathrm{false}$\;
			\EndIf
			\Comment{It holds that $\mathcal B(\phi_1) = \mathcal B(\phi_2)$}\;
			\ForEach{$b \in \mathcal B(\phi_1)$}
				\Comment{In ascending order of $\mathcal F(b)$}\;
				\If{$\phi_1(b) \neq \phi_2(b)$}
					\State \Return $\mathrm{false}$\;
				\EndIf
				\State $C_b \gets$ Possible swaps to obtain $\phi_1^{\succ}(b) = \phi_2^{\succ}(b)$\; \label{line:swap}
			\EndForEach
			\If{there exists a swap in $\bigcap_{b \in \mathcal B(\phi_1)} C_b$ s.t.\ $\phi_1 = \phi_2$}
				\State \Return $\mathrm{true}$\;
			\Else
				\State \Return $\mathrm{false}$\;
			\EndIf
		\end{algorithmic}
	\end{algorithm}
	First, \ac{deft} ensures that $\phi_1$ and $\phi_2$ are defined over the same function domain, i.e., that they have the same number of arguments and that their arguments entail the same set of buckets.
	Thereafter, \ac{deft} iterates over the buckets (which are identical for $\phi_1$ and $\phi_2$) in ascending order of their degree of freedom and ensures that they are mapped to the same multiset of values by $\phi_1$ and $\phi_2$.
	If this initial check fails, $\phi_1$ and $\phi_2$ are not exchangeable and \ac{deft} stops.
	Otherwise, \ac{deft} checks for each bucket $b$ whether the arguments of $\phi_2$ can be rearranged such that $\phi_1^{\succ}(b) = \phi_2^{\succ}(b)$.
	More specifically, for each position $p \in \{1, \dots, \abs{\phi_2^{\succ}(b)}\}$ in $\phi_2^{\succ}(b)$, \ac{deft} looks up all positions $p'$ in $\phi_1^{\succ}(b)$ that contain the same potential value as $\phi_2^{\succ}(b)$ at position $p$.
	Having found all positions $p'$, \ac{deft} knows that the potential values at position $p$ and $p'$ in $\phi_2^{\succ}(b)$ are candidates for swapping to obtain $\phi_1^{\succ}(b) = \phi_2^{\succ}(b)$.
	\Ac{deft} then looks up the assignments (rows in the table) of $\phi_2$ that correspond to the potential values at positions $p$ and $p'$ and builds a dictionary $C_b$ of possible rearrangements for $\phi_2$'s arguments.
	In particular, $C_b$ is a dictionary that maps each argument position $i \in \{1, \dots, m\}$ to a set of possible new argument positions at which the argument $R_i$ can be placed.
	Let $\mathcal A = (a_1, \dots, a_m)$ and $\mathcal A' = (a'_1, \dots, a'_m)$ denote the assignments that $\phi_2$ maps to the potential values at positions $p$ and $p'$ in $\phi_2^{\succ}(b)$.
	To rearrange the order of potential values in $\phi_2^{\succ}(b)$ such that $\phi_1^{\succ}(b) = \phi_2^{\succ}(b)$ holds, the arguments of $\phi_2$ can be rearranged in any way such that $\mathcal A$ maps to $\varphi'$ and $\mathcal A'$ maps to $\varphi$ afterwards.
	Therefore, \ac{deft} iterates over the assignment $\mathcal A$ and stores an entry in $C_b$ for each position $i \in \{1, \dots, m\}$ containing a set of possible rearrangements $\{p_1, \dots, p_{\ell}\}$ such that $a_i = a'_{p_j}$ holds for all $j \in \{1, \dots, \ell\}$.
	Finally, \ac{deft} builds the intersection over the possible rearrangements of all positions in all buckets and checks whether there is a rearrangement left such that the tables of $\phi_1$ and $\phi_2$ are identical.

	\begin{example}
		Take again a look at \cref{fig:example_deg_freedom} and let $b = [2,1]$.
		\Ac{deft} begins with position $p = 1$ in $\phi_2^{\succ}(b)$, which is assigned the value $\varphi_3$.
		The value $\varphi_3$ is located at position $p' = 2$ in $\phi_1^{\succ}(b)$ and hence, \ac{deft} considers the assignments $\mathcal A = (\mathrm{true}, \mathrm{true}, \mathrm{false})$ and $\mathcal A' = (\mathrm{true}, \mathrm{false}, \mathrm{true})$ that belong to the potential values at positions $p = 1$ and $p' = 2$ in $\phi_2^{\succ}(b)$.
		The first position in $\mathcal A$ is assigned the value $\mathrm{true}$.
		In $\mathcal A'$, $\mathrm{true}$ is assigned to the first and third position and thus, position $1$ can be rearranged either at position $1$ or $3$, denoted as $1 \mapsto \{1,3\}$.
		\Ac{deft} continues this step for the remaining values in $\mathcal A$ and obtains $1 \mapsto \{1,3\}$, $2 \mapsto \{1,3\}$, and $3 \mapsto \{2\}$ for $C_b$ after handling position $p = 1$ in $\phi_2^{\succ}(b)$.
		The whole procedure is then repeated for the remaining positions in $\phi_2^{\succ}(b)$ and afterwards for the remaining buckets.
	\end{example}

	The reason we iterate the buckets in ascending order of their degree of freedom is that we can already build the intersection of the sets in $C_b$ after each iteration and immediately stop if the intersection becomes empty for at least one argument position.
	Further implementation details and a more comprehensive example can be found in \cref{appendix:implementation_details}.
	We next validate the practical efficiency of \ac{deft} empirically.

\section{Experiments} \label{sec:deft_experiments}
	We compare the practical performance of \ac{deft} to the \enquote{naive} approach (i.e., iterating over all possible permutations) and the state of the art incorporated in the \ac{cpr} algorithm (which uses buckets as a filtering condition before iterating over permutations, see \cref{appendix:acp_description} for more details).
	For our experiments, we generate factors with $n = 2,4,6,8,10,12,14,16$ Boolean arguments and a proportion $p = 0.0, 0.1, 0.2, 0.5, 0.8, 0.9, 1.0$ of identical potentials.
	For each scenario, we generate exchangeable factors and non-exchangeable factors.
	The exchangeable factors are generated by creating two factors with identical tables and then randomly permuting the arguments of one factor (and rearranging its table accordingly to keep its semantics).

	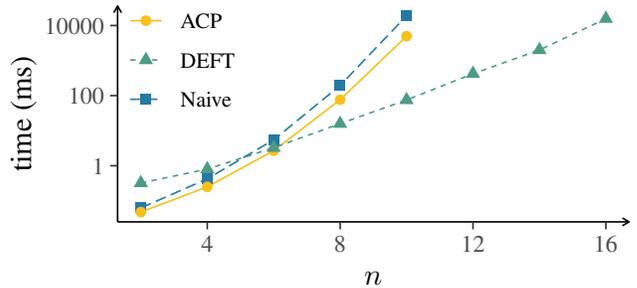
\begin{figure}[t]
		\centering
		\resizebox{\linewidth}{!}{
			\begin{tikzpicture}[x=1pt,y=1pt]
				\definecolor{fillColor}{RGB}{255,255,255}
				\begin{scope}
				\definecolor{drawColor}{RGB}{255,255,255}
				\definecolor{fillColor}{RGB}{255,255,255}
				\end{scope}
				\begin{scope}
				\definecolor{fillColor}{RGB}{255,255,255}

				\definecolor{drawColor}{RGB}{247,192,26}
				
				\path[draw=drawColor,line width= 0.6pt,line join=round] ( 53.46, 34.30) --
					( 77.88, 43.58) --
					(102.31, 56.89) --
					(126.74, 75.65) --
					(151.16, 99.02);
				\definecolor{drawColor}{RGB}{78,155,133}
				
				\path[draw=drawColor,line width= 0.6pt,dash pattern=on 2pt off 2pt ,line join=round] ( 53.46, 45.00) --
					( 77.88, 50.11) --
					(102.31, 57.98) --
					(126.74, 66.83) --
					(151.16, 75.55) --
					(175.59, 85.19) --
					(200.02, 93.95) --
					(224.44,105.46);
				\definecolor{drawColor}{RGB}{37,122,164}
				
				\path[draw=drawColor,line width= 0.6pt,dash pattern=on 4pt off 2pt ,line join=round] ( 53.46, 35.81) --
					( 77.88, 46.52) --
					(102.31, 60.89) --
					(126.74, 80.87) --
					(151.16,106.52);
				\definecolor{fillColor}{RGB}{37,122,164}
				
				\path[fill=fillColor] ( 75.92, 44.56) --
					( 79.85, 44.56) --
					( 79.85, 48.48) --
					( 75.92, 48.48) --
					cycle;
				
				\path[fill=fillColor] (149.20,104.56) --
					(153.13,104.56) --
					(153.13,108.48) --
					(149.20,108.48) --
					cycle;
				
				\path[fill=fillColor] ( 51.50, 33.85) --
					( 55.42, 33.85) --
					( 55.42, 37.78) --
					( 51.50, 37.78) --
					cycle;
				
				\path[fill=fillColor] (100.35, 58.93) --
					(104.27, 58.93) --
					(104.27, 62.85) --
					(100.35, 62.85) --
					cycle;
				
				\path[fill=fillColor] (124.77, 78.91) --
					(128.70, 78.91) --
					(128.70, 82.83) --
					(124.77, 82.83) --
					cycle;
				\definecolor{fillColor}{RGB}{247,192,26}
				
				\path[fill=fillColor] ( 77.88, 43.58) circle (  1.96);
				
				\path[fill=fillColor] (151.16, 99.02) circle (  1.96);
				
				\path[fill=fillColor] ( 53.46, 34.30) circle (  1.96);
				
				\path[fill=fillColor] (102.31, 56.89) circle (  1.96);
				
				\path[fill=fillColor] (126.74, 75.65) circle (  1.96);
				\definecolor{fillColor}{RGB}{78,155,133}
				
				\path[fill=fillColor] (200.02, 97.00) --
					(202.66, 92.43) --
					(197.37, 92.43) --
					cycle;
				
				\path[fill=fillColor] ( 77.88, 53.16) --
					( 80.53, 48.59) --
					( 75.24, 48.59) --
					cycle;
				
				\path[fill=fillColor] (175.59, 88.25) --
					(178.23, 83.67) --
					(172.95, 83.67) --
					cycle;
				
				\path[fill=fillColor] (151.16, 78.60) --
					(153.81, 74.02) --
					(148.52, 74.02) --
					cycle;
				
				\path[fill=fillColor] ( 53.46, 48.05) --
					( 56.10, 43.47) --
					( 50.82, 43.47) --
					cycle;
				
				\path[fill=fillColor] (102.31, 61.04) --
					(104.95, 56.46) --
					( 99.67, 56.46) --
					cycle;
				
				\path[fill=fillColor] (224.44,108.51) --
					(227.08,103.93) --
					(221.80,103.93) --
					cycle;
				
				\path[fill=fillColor] (126.74, 69.88) --
					(129.38, 65.31) --
					(124.09, 65.31) --
					cycle;
				\end{scope}
				\begin{scope}
				\definecolor{drawColor}{RGB}{0,0,0}
				
				\path[draw=drawColor,line width= 0.6pt,line join=round] ( 44.91, 30.69) --
					( 44.91,110.13);
				
				\path[draw=drawColor,line width= 0.6pt,line join=round] ( 46.33,107.67) --
					( 44.91,110.13) --
					( 43.49,107.67);
				\end{scope}
				\begin{scope}
				\definecolor{drawColor}{gray}{0.30}
				
				\node[text=drawColor,anchor=base east,inner sep=0pt, outer sep=0pt, scale=  0.88] at ( 39.96, 48.32) {1};
				
				\node[text=drawColor,anchor=base east,inner sep=0pt, outer sep=0pt, scale=  0.88] at ( 39.96, 74.12) {100};
				
				\node[text=drawColor,anchor=base east,inner sep=0pt, outer sep=0pt, scale=  0.88] at ( 39.96, 99.91) {10000};
				\end{scope}
				\begin{scope}
				\definecolor{drawColor}{gray}{0.20}
				
				\path[draw=drawColor,line width= 0.6pt,line join=round] ( 42.16, 51.35) --
					( 44.91, 51.35);
				
				\path[draw=drawColor,line width= 0.6pt,line join=round] ( 42.16, 77.15) --
					( 44.91, 77.15);
				
				\path[draw=drawColor,line width= 0.6pt,line join=round] ( 42.16,102.95) --
					( 44.91,102.95);
				\end{scope}
				\begin{scope}
				\definecolor{drawColor}{RGB}{0,0,0}
				
				\path[draw=drawColor,line width= 0.6pt,line join=round] ( 44.91, 30.69) --
					(232.99, 30.69);
				
				\path[draw=drawColor,line width= 0.6pt,line join=round] (230.53, 29.26) --
					(232.99, 30.69) --
					(230.53, 32.11);
				\end{scope}
				\begin{scope}
				\definecolor{drawColor}{gray}{0.20}
				
				\path[draw=drawColor,line width= 0.6pt,line join=round] ( 77.88, 27.94) --
					( 77.88, 30.69);
				
				\path[draw=drawColor,line width= 0.6pt,line join=round] (126.74, 27.94) --
					(126.74, 30.69);
				
				\path[draw=drawColor,line width= 0.6pt,line join=round] (175.59, 27.94) --
					(175.59, 30.69);
				
				\path[draw=drawColor,line width= 0.6pt,line join=round] (224.44, 27.94) --
					(224.44, 30.69);
				\end{scope}
				\begin{scope}
				\definecolor{drawColor}{gray}{0.30}
				
				\node[text=drawColor,anchor=base,inner sep=0pt, outer sep=0pt, scale=  0.88] at ( 77.88, 19.68) {4};
				
				\node[text=drawColor,anchor=base,inner sep=0pt, outer sep=0pt, scale=  0.88] at (126.74, 19.68) {8};
				
				\node[text=drawColor,anchor=base,inner sep=0pt, outer sep=0pt, scale=  0.88] at (175.59, 19.68) {12};
				
				\node[text=drawColor,anchor=base,inner sep=0pt, outer sep=0pt, scale=  0.88] at (224.44, 19.68) {16};
				\end{scope}
				\begin{scope}
				\definecolor{drawColor}{RGB}{0,0,0}
				
				\node[text=drawColor,anchor=base,inner sep=0pt, outer sep=0pt, scale=  1.10] at (138.95,  7.64) {$n$};
				\end{scope}
				\begin{scope}
				\definecolor{drawColor}{RGB}{0,0,0}
				
				\node[text=drawColor,rotate= 90.00,anchor=base,inner sep=0pt, outer sep=0pt, scale=  1.10] at ( 13.08, 70.41) {time (ms)};
				\end{scope}
				\begin{scope}
				
				\end{scope}
				\begin{scope}
				\definecolor{drawColor}{RGB}{247,192,26}
				
				\path[draw=drawColor,line width= 0.6pt,line join=round] ( 49.55,104.72) -- ( 61.12,104.72);
				\end{scope}
				\begin{scope}
				\definecolor{fillColor}{RGB}{247,192,26}
				
				\path[fill=fillColor] ( 55.33,104.72) circle (  1.96);
				\end{scope}
				\begin{scope}
				\definecolor{drawColor}{RGB}{78,155,133}
				
				\path[draw=drawColor,line width= 0.6pt,dash pattern=on 2pt off 2pt ,line join=round] ( 49.55, 90.27) -- ( 61.12, 90.27);
				\end{scope}
				\begin{scope}
				\definecolor{fillColor}{RGB}{78,155,133}
				
				\path[fill=fillColor] ( 55.33, 93.32) --
					( 57.98, 88.74) --
					( 52.69, 88.74) --
					cycle;
				\end{scope}
				\begin{scope}
				\definecolor{drawColor}{RGB}{37,122,164}
				
				\path[draw=drawColor,line width= 0.6pt,dash pattern=on 4pt off 2pt ,line join=round] ( 49.55, 75.82) -- ( 61.12, 75.82);
				\end{scope}
				\begin{scope}
				\definecolor{fillColor}{RGB}{37,122,164}
				
				\path[fill=fillColor] ( 53.37, 73.85) --
					( 57.30, 73.85) --
					( 57.30, 77.78) --
					( 53.37, 77.78) --
					cycle;
				\end{scope}
				\begin{scope}
				\definecolor{drawColor}{RGB}{0,0,0}
				
				\node[text=drawColor,anchor=base west,inner sep=0pt, outer sep=0pt, scale=  0.80] at ( 68.06,101.97) {ACP};
				\end{scope}
				\begin{scope}
				\definecolor{drawColor}{RGB}{0,0,0}
				
				\node[text=drawColor,anchor=base west,inner sep=0pt, outer sep=0pt, scale=  0.80] at ( 68.06, 87.52) {DEFT};
				\end{scope}
				\begin{scope}
				\definecolor{drawColor}{RGB}{0,0,0}
				
				\node[text=drawColor,anchor=base west,inner sep=0pt, outer sep=0pt, scale=  0.80] at ( 68.06, 73.06) {Naive};
				\end{scope}
			\end{tikzpicture}
		}
		\caption{Average run times of the \enquote{naive} approach, the approach used in \ac{cpr}, and \ac{deft} for different numbers of arguments $n$. For each choice of $n$, the proportion of identical potentials is varied between $0.0$ and $1.0$ and both exchangeable as well as non-exchangeable factors are considered.}
		\label{fig:experiments}
	\end{figure}

	We run each algorithm with a timeout of 30 minutes per instance and report the average run time over all instances for each choice of $n$.
	\Cref{fig:experiments} displays the average run times on a logarithmic scale.
	While both the naive approach and \ac{cpr} are fast on small instances with up to $n = 8$ arguments, they do not scale for larger instances.
	After $n = 10$, both the naive approach and \ac{cpr} run into timeouts, which is not surprising as they both have to iterate over $O(n!)$ permutations.
	\Ac{deft}, on the other hand, is able to solve all instances within the specified timeout and is able to handle instances having nearly twice as many arguments as the instances that can be solved by the naive approach and \ac{cpr}.
	For small values of $n$, \ac{deft} is slightly slower than the naive approach and \ac{cpr} due to additional pre-processing and with increasing $n$, \ac{deft} greatly benefits from its pre-processing.
	In particular, \ac{deft} solves instances with $n = 16$ in about ten seconds on average while the naive approach and \ac{cpr} are in general not able to solve instances with $n = 12$ within 30 minutes.
	Finally, we remark that \ac{cpr} is able to solve instances with $n > 10$ that are not exchangeable within the specified timeout but as \ac{cpr} does not solve any instance of exchangeable factors with $n > 10$ within the timeout, it is not possible to compute a meaningful average run time.
	We thus provide further experimental results in \cref{appendix:further_experiments}, where we give additional results specifically for individual scenarios.

\section{Conclusion}
	We introduce the \ac{deft} algorithm to efficiently detect exchangeable factors in \acp{fg}.
	\Ac{deft} uses buckets to drastically reduce the required computational effort in many practical settings.
	In particular, we prove that the number of table comparisons needed to check whether two factors are exchangeable is upper-bounded by the number of identical potential values within the buckets of the factors.
	By exploiting this upper bound, \ac{deft} is able to significantly reduce the number of permutations that must be considered to check whether two factors are exchangeable.


\section*{Acknowledgements}
This work is mainly funded by the BMBF project AnoMed 16KISA057 and 16KISA050K.
The authors would like to thank Tanya Braun, Ralf Möller, and the anonymous reviewers for their valuable comments.

\bibliographystyle{flairs}
\bibliography{references.bib}

\clearpage
\appendix
\acresetall

\section{Missing Proofs}
\upperboundCorollary*
\begin{proof}
	Let $\{R'_1, \dots, R'_k\}$ denote the largest subset of $\phi$'s arguments such that all arguments in $\{R'_1, \dots, R'_k\}$ have the same range $\mathcal V$, i.e., $\mathcal V = \mathcal R(R'_1) = \dots = \mathcal R(R'_k)$.
	If $k = 1$, $\mathcal F(\phi) = 1$ and thus our claim holds.
	Therefore, we assume that $k > 1$ for the remaining part of this proof.
	Then, there exists a bucket $b = [n_1, \dots, n_{\abs{\mathcal V}}]$ (with $\sum_i n_i = k$) such that one of the $n_i$ is set to $k-1$, another one is set to $1$, and all remaining are set to $0$.
	In other words, bucket $b$ represents all assignments such that there is a $v_i \in \mathcal V$ appearing $k-1$ times and there is a $v_j \in \mathcal V$ with $v_j \neq v_i$ appearing once.
	Therefore, $b$ represents exactly $k$ assignments because there are $k$ possible positions for $v_i$ to be located at.
	Consequently, $\abs{\phi(b)} = k$ because there is a potential in $b$ for each assignment represented by $b$.
	As $\mathcal F(b)$ is maximal when all potentials in $\phi(b)$ are identical, it holds that $\mathcal F(b) \leq k!$.
	Finally, we have $k \leq n$ (because $\phi$ has $n$ arguments) and as $\mathcal F(\phi) = \min_{b \in \{b \mid b \in \mathcal B(\phi) \land \abs{\phi(b)} > 1\}} \mathcal F(b)$, it holds that $\mathcal F(\phi) \leq \mathcal F(b) \leq k! \leq n!$.
\end{proof}

\section{Formal Description of the Advanced Colour Passing Algorithm} \label{appendix:acp_description}
The \ac{cpr} algorithm introduced by \citeauthor{Luttermann2024a}~(\citeyear{Luttermann2024a}) builds on the \acl{cp} algorithm~\cite{Kersting2009a,Ahmadi2013a} and solves the problem of constructing a lifted representation (i.e., a so-called \acl{pfg}) from a given \ac{fg}.
The idea of \ac{cpr} is to first find symmetries in a propositional \ac{fg} and then group together symmetric subgraphs.
\Ac{cpr} looks for symmetries based on potentials of factors, on ranges and evidence of \acp{rv}, as well as on the graph structure by passing around colours.
\Cref{alg:cp_revisited} provides a formal description of the \ac{cpr} algorithm, which proceeds as follows.

\begin{algorithm}[t]
	\caption{Advanced Colour Passing (as introduced by \citeauthor{Luttermann2024a}, \citeyear{Luttermann2024a})}
	\label{alg:cp_revisited}
	\alginput{An \ac{fg} $G$ with \acp{rv} $\boldsymbol R = \{R_1, \ldots, R_n\}$, and factors $\boldsymbol \Phi = \{\phi_1, \ldots, \phi_m\}$, as well as a set of evidence $\boldsymbol E = \{R_1 = r_1, \ldots, R_k = r_k\}$.} \\
	\algoutput{A lifted representation $G'$ in form of a \ac{pfg} with equivalent semantics to $G$.}
	\begin{algorithmic}[1]
		\State Assign each $R_i$ a colour according to $\mathcal R(R_i)$ and $\boldsymbol E$\;
		\State Assign each $\phi_i$ a colour according to order-independent potentials and rearrange arguments accordingly\;
		\Repeat
			\For{each factor $\phi \in \boldsymbol \Phi$}
				\State $signature_{\phi} \gets [\,]$\;
				\For{each \ac{rv} $R \in neighbours(G, \phi)$}
					\State\Comment{In order of appearance in $\phi$}\;
					\State $append(signature_{\phi}, R.colour)$\;
				\EndFor
				\State $append(signature_{\phi}, \phi.colour)$\;
			\EndFor
			\State Group together all $\phi$s with the same signature\;
			\State Assign each such cluster a unique colour\;
			\State Set $\phi.colour$ correspondingly for all $\phi$s\;
			\For{each \ac{rv} $R \in \boldsymbol R$}
				\State $signature_{R} \gets [\,]$\;
				\For{each factor $\phi \in neighbours(G, R)$}
					\If{$\phi$ is commutative w.r.t.\ $\boldsymbol S$ and $R \in \boldsymbol S$}
						\State $append(signature_{R}, (\phi.colour, 0))$\;
					\Else
						\State $append(signature_{R}, (\phi.colour, p(R, \phi)))$\;
					\EndIf
				\EndFor
				\State Sort $signature_{R}$ according to colour\;
				\State $append(signature_{R}, R.colour)$\;
			\EndFor
			\State Group together all $R$s with the same signature\;
			\State Assign each such cluster a unique colour\;
			\State Set $R.colour$ correspondingly for all $R$s\;
		\Until{grouping does not change}
		\State $G' \gets$ construct \acs{pfg} from groupings\;
	\end{algorithmic}
\end{algorithm}

\Ac{cpr} begins with the colour assignment to variable nodes, meaning that all \acp{rv} that have the same range and observed event are assigned the same colour.
Thereafter, \ac{cpr} assigns colours to factor nodes such that factors representing identical potentials are assigned the same colour.
Two factors represent identical potentials if they are exchangeable according to \cref{def:exchangeable}, i.e., if there exists a rearrangement of one of the factor's arguments such that both factors have identical tables of potentials when comparing them row by row.
In its original form, \ac{cpr} first checks whether two factors entail the same buckets and whether all buckets contain the same potential values.
If this initial check fails, the two factors cannot be exchangeable and \ac{cpr} assigns different colours to them, otherwise, \ac{cpr} iterates over all possible permutations of the arguments of one factor to check whether there exists a permutation such that the tables of potential mappings are identical for both factors.
The \ac{deft} algorithm avoids this exhaustive search over all permutations of arguments by restricting the search space to a small subset of possible permutations.
After the initial colour assignments, \ac{cpr} passes the colours around.
\Ac{cpr} first passes the colours from every variable node to its neighbouring factor nodes and afterwards, every factor node $\phi$ sends its colour plus the position $p(R, \phi)$ of $R$ in $\phi$'s argument list to all of its neighbouring variable nodes $R$.
For more details about the colour passing procedure and the grouping of nodes, we refer the reader to \cite{Luttermann2024a}.
The authors also provide extensive results about the benefits of constructing a lifted representation for probabilistic inference.

\section{Implementation Details of \acs{deft}} \label{appendix:implementation_details}
To complement the description of the \ac{deft} algorithm given in \cref{alg:deft}, we provide implementation details on the computation of possible swaps (rearrangements) of $\phi_2$'s arguments to obtain identically ordered buckets for $\phi_1(R_1, \dots, R_n)$ and $\phi_2(R'_1, \dots, R'_m)$ (\cref{line:swap} in \cref{alg:deft}) in this section.
To check whether it is possible to rearrange the arguments of $\phi_2$ such that $\phi_1^{\succ}(b) = \phi_2^{\succ}(b)$ holds for a bucket $b$, \ac{deft} proceeds as follows.
For each position $p \in \{1, \dots, \abs{\phi_2^{\succ}(b)}\}$ in $\phi_2^{\succ}(b)$, \ac{deft} looks up all positions $p'$ in $\phi_1^{\succ}(b)$ that contain the same potential value as $\phi_2^{\succ}(b)$ at position $p$ and then builds sets of possible rearrangements of arguments for each position.
Let $\varphi$ and $\varphi'$ denote the two potential values at positions $p$ and $p'$ in $\phi_2^{\succ}(b)$, respectively, which in turn correspond to two assignments (rows) in the table of $\phi_2$.
Note that it is also possible that the corresponding assignments are identical (i.e., they refer to the same row in the table of $\phi_2$).
Let $\mathcal A = (a_1, \dots, a_m)$ and $\mathcal A' = (a'_1, \dots, a'_m)$ denote the assignments that are mapped to $\varphi$ and $\varphi'$, respectively, by $\phi_2$.
To swap the positions of $\varphi$ and $\varphi'$ in $\phi_2^{\succ}(b)$ (i.e., to rearrange the order of potential values in $\phi_2^{\succ}(b)$ such that $\phi_1^{\succ}(b) = \phi_2^{\succ}(b)$), the arguments of $\phi_2$ can be rearranged in any way such that $\mathcal A$ maps to $\varphi'$ and $\mathcal A'$ maps to $\varphi$ afterwards.
Therefore, \ac{deft} iterates over the assignment $\mathcal A$ and builds sets of possible rearrangements for each position $1, \dots, m$ in $\mathcal A$.
More specifically, \ac{deft} stores for each position $i \in \{1, \dots, m\}$ a set of positions $\{p_1, \dots, p_{\ell}\}$ such that $a_i = a'_{p_j}$ for all $j \in \{1, \dots, \ell\}$.

\begin{example} \label{ex:swap_implementation_1}
	Consider again \cref{fig:example_deg_freedom} and let $b = [2,1]$.
	\Ac{deft} begins by considering the position $p = 1$ in $\phi_2^{\succ}(b)$, which is assigned the value $\varphi_3$.
	Then, \ac{deft} looks up all positions in $\phi_1^{\succ}(b)$ that are assigned the value $\varphi_3$ and obtains a single position $p' = 2$.
	$\phi_2^{\succ}(b)$ contains the value $\varphi_5$ at position $p' = 2$.
	The corresponding assignments of $\varphi_3$ and $\varphi_5$ in the table of $\phi_2$ are given by the assignments $\mathcal A = (\mathrm{true}, \mathrm{true}, \mathrm{false})$ and $\mathcal A' = (\mathrm{true}, \mathrm{false}, \mathrm{true})$, respectively.
	Then, \ac{deft} builds sets of possible rearrangements of $\phi_2$'s arguments as follows.
	The first position in the argument list is assigned the value $\mathrm{true}$ in $\mathcal A$.
	In $\mathcal A'$, $\mathrm{true}$ is assigned to the first and third position, i.e., position one can be rearranged either at position one or three, denoted as $1 \mapsto \{1,3\}$.
	\Ac{deft} continues this step for the remaining values in $\mathcal A$ and obtains $1 \mapsto \{1,3\}$, $2 \mapsto \{1,3\}$, and $3 \mapsto \{2\}$ for position $p = 1$ in $\phi_2^{\succ}(b)$.

	Next, \ac{deft} considers the position $p = 2$ in $\phi_2^{\succ}(b)$, which is assigned the value $\varphi_5$ and looks up the position $p' = 3$ of $\varphi_5$ in $\phi_1^{\succ}(b)$.
	$\phi_2^{\succ}(b)$ contains the value $\varphi_2$ at position $p' = 3$.
	The corresponding assignments of $\varphi_5$ and $\varphi_2$ are given by $\mathcal A = (\mathrm{true}, \mathrm{false}, \mathrm{true})$ and $\mathcal A' = (\mathrm{false}, \mathrm{true}, \mathrm{true})$, respectively.
	Therefore, \ac{deft} obtains $1 \mapsto \{2,3\}$, $2 \mapsto \{1\}$, and $3 \mapsto \{2,3\}$ as possible rearrangements of $\phi_2$'s arguments for position $p = 2$ in $\phi_2^{\succ}(b)$.

	\Ac{deft} then repeats the procedure for the last position $p = 3$.
	$\phi_2^{\succ}(b)$ contains the value $\varphi_2$ at position $p = 3$ and \ac{deft} obtains $p' = 1$ as $\varphi_2$ occurs at position $p' = 1$ in $\phi_1^{\succ}(b)$.
	$\phi_2^{\succ}(b)$ contains the value $\varphi_3$ at position $p' = 1$.
	The corresponding assignments of $\varphi_2$ and $\varphi_3$ are given by $\mathcal A = (\mathrm{false}, \mathrm{true}, \mathrm{true})$ and $\mathcal A' = (\mathrm{true}, \mathrm{true}, \mathrm{false})$, respectively.
	In consequence, \ac{deft} obtains the possible rearrangements $1 \mapsto \{3\}$, $2 \mapsto \{1,2\}$, and $3 \mapsto \{1,2\}$ for position $p = 3$ in $\phi_2^{\succ}(b)$.

	As every position $p \in \{1, 2, 3\}$ must be arranged such that the potential value located at position $p$ in $\phi_1^{\succ}(b)$ is identical to the potential value located at position $p$ in $\phi_2^{\succ}(b)$, \ac{deft} computes the intersection of all sets of possible rearrangements for each position to obtain a possible rearrangement of $\phi_2$'s arguments that ensures $\phi_1^{\succ}(b) = \phi_2^{\succ}(b)$.
	Building the intersections for the sets of possible rearrangements, for position one it holds that $1 \mapsto \{1,3\} \cap \{2,3\} \cap \{3\} = \{3\}$, for position two we have $2 \mapsto \{1,3\} \cap \{1\} \cap \{1,3\} = \{1\}$, and for position three we obtain $3 \mapsto \{2\} \cap \{2,3\} \cap \{1,2\} = \{2\}$.
\end{example}

As we have seen, all potential values must be located at the same position in their respective bucket for both $\phi_1$ and $\phi_2$ and in order to ensure this, \ac{deft} computes the intersection of all sets of possible rearrangements for each position in the arguments of $\phi_2$.
Consequently, if there exists an argument position $i$ such that $i \mapsto \emptyset$ after the intersection, there is no possibility to rearrange the arguments of $\phi_2$ to ensure that $\phi_1^{\succ}(b) = \phi_2^{\succ}(b)$ holds and \ac{deft} immediately stops.
Further, as $\phi_1^{\succ}(b) = \phi_2^{\succ}(b)$ must hold for \emph{all} buckets $b \in \mathcal B(\phi_1) = \mathcal B(\phi_2)$, \ac{deft} builds sets of possible rearrangements for each bucket and then computes the intersection of them.
If there is an argument position for which the intersection is empty, no rearrangement is possible to ensure that $\phi_1^{\succ}(b) = \phi_2^{\succ}(b)$ holds for all buckets.

\begin{example}
	To continue \cref{ex:swap_implementation_1}, \ac{deft} computes the intersection of the sets of possible rearrangements for all remaining buckets $b \in \{[3,0],[0,3],[1,2]\}$.
	Afterwards, \ac{deft} obtains the possible rearrangements $1 \mapsto \{3\}$, $2 \mapsto \{1\}$, and $3 \mapsto \{2\}$.
	Finally, \ac{deft} rearranges the arguments of $\phi_2$ such that position one ($R_4$) is located at position three, position two ($R_5$) is located at position one, and position three ($R_6$) is located at position two, i.e., we obtain $R_5$, $R_6$, $R_4$ as the new argument list for $\phi_2$, and \ac{deft} verifies that the tables of $\phi_1$ and $\phi_2$ are identical after this rearrangement.
\end{example}

Moreover, we remark that it is also conceivable to compute the intersection of possible rearrangements over a subset of the buckets only.
In particular, each bucket further restricts the set of possible rearrangements, that is, the less buckets we consider, the more possible rearrangements are permitted, which, in turn, must all be considered for verification to determine whether $\phi_1$ and $\phi_2$ are exchangeable.

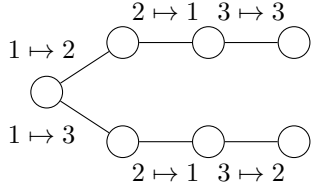
\begin{figure}[t]
	\centering
	\begin{tikzpicture}
		\node[circle, draw, minimum size = 1.2em] (1) {};
		\node[circle, draw, minimum size = 1.2em, above right = 1.0em and 2.0em of 1] (2l) {};
		\node[circle, draw, minimum size = 1.2em, below right = 1.0em and 2.0em of 1] (2r) {};
		\node[circle, draw, minimum size = 1.2em, right = 2.0em of 2l] (3l) {};
		\node[circle, draw, minimum size = 1.2em, right = 2.0em of 2r] (3r) {};
		\node[circle, draw, minimum size = 1.2em, right = 2.0em of 3l] (4l) {};
		\node[circle, draw, minimum size = 1.2em, right = 2.0em of 3r] (4r) {};
	
		\draw (1) edge node[above left] {$1 \mapsto 2$} (2l);
		\draw (1) edge node[below left] {$1 \mapsto 3$} (2r);
		\draw (2l) edge node[above, yshift = 0.5em] {$2 \mapsto 1$} (3l);
		\draw (2r) edge node[below, yshift = -0.5em] {$2 \mapsto 1$} (3r);
		\draw (3l) edge node[above, yshift = 0.5em] {$3 \mapsto 3$} (4l);
		\draw (3r) edge node[below, yshift = -0.5em] {$3 \mapsto 2$} (4r);
	\end{tikzpicture}
	\caption{Possible rearrangements of the argument positions $1$, $2$, and $3$ for the sets of possible rearrangements $1 \mapsto \{2,3\}$, $2 \mapsto \{1\}$, and $3 \mapsto \{2,3\}$.}
	\label{fig:example_permtree}
\end{figure}

\begin{example}
	Consider again \cref{fig:example_deg_freedom} and let $b = [1,2]$.
	After iterating over all positions in $\phi_2^{\succ}(b)$, \ac{deft} obtains the sets of possible rearrangements $1 \mapsto \{2,3\}$, $2 \mapsto \{1\}$, and $3 \mapsto \{2,3\}$.
	Consequently, there are multiple options for rearranging the arguments of $\phi_2$ if we consider only the bucket $b$.
	In case there are multiple options for rearranging the arguments of $\phi_2$, \ac{deft} checks whether there is an option such that the tables of $\phi_1$ and $\phi_2$ are identical.
	The possible rearrangements of $\phi_2$'s arguments are illustrated in \cref{fig:example_permtree}.
	\Ac{deft} tries them in any order, so it might start with placing $R_4$ (currently position $1$) at position $2$, then places $R_5$ (currently position $2$) at position $1$, and finally places $R_6$ (currently position $3$) at position $3$.
	Note that we originally had $3 \mapsto \{2,3\}$, i.e., $R_6$ could potentially be placed at position $2$ as well but as we already fixed position $2$ (by placing $R_4$ there), the only remaining possible position for $R_6$ is position $3$.
	After the rearrangement to $R_5$, $R_4$, $R_6$, the verification that the tables of $\phi_1$ and $\phi_2$ are identical fails and \ac{deft} tries the next possible rearrangement.
	The next rearrangement is then $R_4$, $R_6$, $R_5$ and the verification succeeds.
	In case none of the possible rearrangements results in identical tables, $\phi_1$ and $\phi_2$ are not exchangeable.
\end{example}

Note that the number of paths in the tree depicted in \cref{fig:example_permtree} depends on the number of identical potential values in the bucket, which is mostly small in practical applications as in general different assignments map to different potential values.
Therefore, \ac{deft} is guaranteed to never consider more rearrangements than previous approaches.

In our implementation, we deploy a heuristic within \ac{deft} such that only the five buckets with the lowest degree of freedom are considered to compute the intersection of possible rearrangements.
The use of such an heuristic does not impact the correctness of \ac{deft} while at the same time significantly reduces the overhead for computing the intersections over large buckets.

\section{Further Experimental Results} \label{appendix:further_experiments}
\begin{figure*}[t]
		\centering
		\begin{tikzpicture}[x=1pt,y=1pt]
			\definecolor{fillColor}{RGB}{255,255,255}
			\path[use as bounding box,fill=fillColor,fill opacity=0.00] (0,0) rectangle (238.49,115.63);
			\begin{scope}
			\path[clip] (  0.00,  0.00) rectangle (238.49,115.63);
			\definecolor{drawColor}{RGB}{255,255,255}
			\definecolor{fillColor}{RGB}{255,255,255}
			
			\path[draw=drawColor,line width= 0.6pt,line join=round,line cap=round,fill=fillColor] (  0.00,  0.00) rectangle (238.49,115.63);
			\end{scope}
			\begin{scope}
			\path[clip] ( 44.91, 30.69) rectangle (232.99,110.13);
			\definecolor{fillColor}{RGB}{255,255,255}
			
			\path[fill=fillColor] ( 44.91, 30.69) rectangle (232.99,110.13);
			\definecolor{drawColor}{RGB}{247,192,26}
			
			\path[draw=drawColor,line width= 0.6pt,line join=round] ( 53.46, 34.30) --
				( 77.88, 41.03) --
				(102.31, 48.68) --
				(126.74, 56.74) --
				(151.16, 64.57) --
				(175.59, 72.37) --
				(200.02, 80.50) --
				(224.44, 88.87);
			\definecolor{drawColor}{RGB}{78,155,133}
			
			\path[draw=drawColor,line width= 0.6pt,dash pattern=on 2pt off 2pt ,line join=round] ( 53.46, 43.01) --
				( 77.88, 51.26) --
				(102.31, 59.79) --
				(126.74, 68.07) --
				(151.16, 76.33) --
				(175.59, 84.87) --
				(200.02, 94.87) --
				(224.44,106.52);
			\definecolor{drawColor}{RGB}{37,122,164}
			
			\path[draw=drawColor,line width= 0.6pt,dash pattern=on 4pt off 2pt ,line join=round] ( 53.46, 41.56) --
				( 77.88, 51.16) --
				(102.31, 62.87) --
				(126.74, 81.13) --
				(151.16,104.39);
			\definecolor{fillColor}{RGB}{37,122,164}
			
			\path[fill=fillColor] ( 75.92, 49.20) --
				( 79.85, 49.20) --
				( 79.85, 53.12) --
				( 75.92, 53.12) --
				cycle;
			
			\path[fill=fillColor] (149.20,102.42) --
				(153.13,102.42) --
				(153.13,106.35) --
				(149.20,106.35) --
				cycle;
			
			\path[fill=fillColor] ( 51.50, 39.59) --
				( 55.42, 39.59) --
				( 55.42, 43.52) --
				( 51.50, 43.52) --
				cycle;
			
			\path[fill=fillColor] (100.35, 60.90) --
				(104.27, 60.90) --
				(104.27, 64.83) --
				(100.35, 64.83) --
				cycle;
			
			\path[fill=fillColor] (124.77, 79.17) --
				(128.70, 79.17) --
				(128.70, 83.09) --
				(124.77, 83.09) --
				cycle;
			\definecolor{fillColor}{RGB}{247,192,26}
			
			\path[fill=fillColor] (200.02, 80.50) circle (  1.96);
			
			\path[fill=fillColor] ( 77.88, 41.03) circle (  1.96);
			
			\path[fill=fillColor] (175.59, 72.37) circle (  1.96);
			
			\path[fill=fillColor] (151.16, 64.57) circle (  1.96);
			
			\path[fill=fillColor] ( 53.46, 34.30) circle (  1.96);
			
			\path[fill=fillColor] (102.31, 48.68) circle (  1.96);
			
			\path[fill=fillColor] (224.44, 88.87) circle (  1.96);
			
			\path[fill=fillColor] (126.74, 56.74) circle (  1.96);
			\definecolor{fillColor}{RGB}{78,155,133}
			
			\path[fill=fillColor] (200.02, 97.93) --
				(202.66, 93.35) --
				(197.37, 93.35) --
				cycle;
			
			\path[fill=fillColor] ( 77.88, 54.31) --
				( 80.53, 49.74) --
				( 75.24, 49.74) --
				cycle;
			
			\path[fill=fillColor] (175.59, 87.92) --
				(178.23, 83.35) --
				(172.95, 83.35) --
				cycle;
			
			\path[fill=fillColor] (151.16, 79.38) --
				(153.81, 74.80) --
				(148.52, 74.80) --
				cycle;
			
			\path[fill=fillColor] ( 53.46, 46.06) --
				( 56.10, 41.49) --
				( 50.82, 41.49) --
				cycle;
			
			\path[fill=fillColor] (102.31, 62.84) --
				(104.95, 58.27) --
				( 99.67, 58.27) --
				cycle;
			
			\path[fill=fillColor] (224.44,109.57) --
				(227.08,105.00) --
				(221.80,105.00) --
				cycle;
			
			\path[fill=fillColor] (126.74, 71.12) --
				(129.38, 66.54) --
				(124.09, 66.54) --
				cycle;
			\end{scope}
			\begin{scope}
			\path[clip] (  0.00,  0.00) rectangle (238.49,115.63);
			\definecolor{drawColor}{RGB}{0,0,0}
			
			\path[draw=drawColor,line width= 0.6pt,line join=round] ( 44.91, 30.69) --
				( 44.91,110.13);
			
			\path[draw=drawColor,line width= 0.6pt,line join=round] ( 46.33,107.67) --
				( 44.91,110.13) --
				( 43.49,107.67);
			\end{scope}
			\begin{scope}
			\path[clip] (  0.00,  0.00) rectangle (238.49,115.63);
			\definecolor{drawColor}{gray}{0.30}
			
			\node[text=drawColor,anchor=base east,inner sep=0pt, outer sep=0pt, scale=  0.88] at ( 39.96, 52.81) {1};
			
			\node[text=drawColor,anchor=base east,inner sep=0pt, outer sep=0pt, scale=  0.88] at ( 39.96, 76.40) {100};
			
			\node[text=drawColor,anchor=base east,inner sep=0pt, outer sep=0pt, scale=  0.88] at ( 39.96, 99.99) {10000};
			\end{scope}
			\begin{scope}
			\path[clip] (  0.00,  0.00) rectangle (238.49,115.63);
			\definecolor{drawColor}{gray}{0.20}
			
			\path[draw=drawColor,line width= 0.6pt,line join=round] ( 42.16, 55.84) --
				( 44.91, 55.84);
			
			\path[draw=drawColor,line width= 0.6pt,line join=round] ( 42.16, 79.43) --
				( 44.91, 79.43);
			
			\path[draw=drawColor,line width= 0.6pt,line join=round] ( 42.16,103.02) --
				( 44.91,103.02);
			\end{scope}
			\begin{scope}
			\path[clip] (  0.00,  0.00) rectangle (238.49,115.63);
			\definecolor{drawColor}{RGB}{0,0,0}
			
			\path[draw=drawColor,line width= 0.6pt,line join=round] ( 44.91, 30.69) --
				(232.99, 30.69);
			
			\path[draw=drawColor,line width= 0.6pt,line join=round] (230.53, 29.26) --
				(232.99, 30.69) --
				(230.53, 32.11);
			\end{scope}
			\begin{scope}
			\path[clip] (  0.00,  0.00) rectangle (238.49,115.63);
			\definecolor{drawColor}{gray}{0.20}
			
			\path[draw=drawColor,line width= 0.6pt,line join=round] ( 77.88, 27.94) --
				( 77.88, 30.69);
			
			\path[draw=drawColor,line width= 0.6pt,line join=round] (126.74, 27.94) --
				(126.74, 30.69);
			
			\path[draw=drawColor,line width= 0.6pt,line join=round] (175.59, 27.94) --
				(175.59, 30.69);
			
			\path[draw=drawColor,line width= 0.6pt,line join=round] (224.44, 27.94) --
				(224.44, 30.69);
			\end{scope}
			\begin{scope}
			\path[clip] (  0.00,  0.00) rectangle (238.49,115.63);
			\definecolor{drawColor}{gray}{0.30}
			
			\node[text=drawColor,anchor=base,inner sep=0pt, outer sep=0pt, scale=  0.88] at ( 77.88, 19.68) {4};
			
			\node[text=drawColor,anchor=base,inner sep=0pt, outer sep=0pt, scale=  0.88] at (126.74, 19.68) {8};
			
			\node[text=drawColor,anchor=base,inner sep=0pt, outer sep=0pt, scale=  0.88] at (175.59, 19.68) {12};
			
			\node[text=drawColor,anchor=base,inner sep=0pt, outer sep=0pt, scale=  0.88] at (224.44, 19.68) {16};
			\end{scope}
			\begin{scope}
			\path[clip] (  0.00,  0.00) rectangle (238.49,115.63);
			\definecolor{drawColor}{RGB}{0,0,0}
			
			\node[text=drawColor,anchor=base,inner sep=0pt, outer sep=0pt, scale=  1.10] at (138.95,  7.64) {$n$};
			\end{scope}
			\begin{scope}
			\path[clip] (  0.00,  0.00) rectangle (238.49,115.63);
			\definecolor{drawColor}{RGB}{0,0,0}
			
			\node[text=drawColor,rotate= 90.00,anchor=base,inner sep=0pt, outer sep=0pt, scale=  1.10] at ( 13.08, 70.41) {time (ms)};
			\end{scope}
			\begin{scope}
			\path[clip] (  0.00,  0.00) rectangle (238.49,115.63);
			
			\path[] ( 42.61, 63.09) rectangle ( 96.11,117.45);
			\end{scope}
			\begin{scope}
			\path[clip] (  0.00,  0.00) rectangle (238.49,115.63);
			\definecolor{drawColor}{RGB}{247,192,26}
			
			\path[draw=drawColor,line width= 0.6pt,line join=round] ( 49.55,104.72) -- ( 61.12,104.72);
			\end{scope}
			\begin{scope}
			\path[clip] (  0.00,  0.00) rectangle (238.49,115.63);
			\definecolor{fillColor}{RGB}{247,192,26}
			
			\path[fill=fillColor] ( 55.33,104.72) circle (  1.96);
			\end{scope}
			\begin{scope}
			\path[clip] (  0.00,  0.00) rectangle (238.49,115.63);
			\definecolor{drawColor}{RGB}{78,155,133}
			
			\path[draw=drawColor,line width= 0.6pt,dash pattern=on 2pt off 2pt ,line join=round] ( 49.55, 90.27) -- ( 61.12, 90.27);
			\end{scope}
			\begin{scope}
			\path[clip] (  0.00,  0.00) rectangle (238.49,115.63);
			\definecolor{fillColor}{RGB}{78,155,133}
			
			\path[fill=fillColor] ( 55.33, 93.32) --
				( 57.98, 88.74) --
				( 52.69, 88.74) --
				cycle;
			\end{scope}
			\begin{scope}
			\path[clip] (  0.00,  0.00) rectangle (238.49,115.63);
			\definecolor{drawColor}{RGB}{37,122,164}
			
			\path[draw=drawColor,line width= 0.6pt,dash pattern=on 4pt off 2pt ,line join=round] ( 49.55, 75.82) -- ( 61.12, 75.82);
			\end{scope}
			\begin{scope}
			\path[clip] (  0.00,  0.00) rectangle (238.49,115.63);
			\definecolor{fillColor}{RGB}{37,122,164}
			
			\path[fill=fillColor] ( 53.37, 73.85) --
				( 57.30, 73.85) --
				( 57.30, 77.78) --
				( 53.37, 77.78) --
				cycle;
			\end{scope}
			\begin{scope}
			\path[clip] (  0.00,  0.00) rectangle (238.49,115.63);
			\definecolor{drawColor}{RGB}{0,0,0}
			
			\node[text=drawColor,anchor=base west,inner sep=0pt, outer sep=0pt, scale=  0.80] at ( 68.06,101.97) {ACP};
			\end{scope}
			\begin{scope}
			\path[clip] (  0.00,  0.00) rectangle (238.49,115.63);
			\definecolor{drawColor}{RGB}{0,0,0}
			
			\node[text=drawColor,anchor=base west,inner sep=0pt, outer sep=0pt, scale=  0.80] at ( 68.06, 87.52) {DEFT};
			\end{scope}
			\begin{scope}
			\path[clip] (  0.00,  0.00) rectangle (238.49,115.63);
			\definecolor{drawColor}{RGB}{0,0,0}
			
			\node[text=drawColor,anchor=base west,inner sep=0pt, outer sep=0pt, scale=  0.80] at ( 68.06, 73.06) {Naive};
			\end{scope}
		\end{tikzpicture}
		\begin{tikzpicture}[x=1pt,y=1pt]
			\definecolor{fillColor}{RGB}{255,255,255}
			\path[use as bounding box,fill=fillColor,fill opacity=0.00] (0,0) rectangle (238.49,115.63);
			\begin{scope}
			\path[clip] (  0.00,  0.00) rectangle (238.49,115.63);
			\definecolor{drawColor}{RGB}{255,255,255}
			\definecolor{fillColor}{RGB}{255,255,255}
			
			\path[draw=drawColor,line width= 0.6pt,line join=round,line cap=round,fill=fillColor] (  0.00,  0.00) rectangle (238.49,115.63);
			\end{scope}
			\begin{scope}
			\path[clip] ( 44.91, 30.69) rectangle (232.99,110.13);
			\definecolor{fillColor}{RGB}{255,255,255}
			
			\path[fill=fillColor] ( 44.91, 30.69) rectangle (232.99,110.13);
			\definecolor{drawColor}{RGB}{247,192,26}
			
			\path[draw=drawColor,line width= 0.6pt,line join=round] ( 53.46, 35.57) --
				( 77.88, 44.89) --
				(102.31, 57.55) --
				(126.74, 76.22) --
				(151.16, 99.58);
			\definecolor{drawColor}{RGB}{78,155,133}
			
			\path[draw=drawColor,line width= 0.6pt,dash pattern=on 2pt off 2pt ,line join=round] ( 53.46, 46.16) --
				( 77.88, 50.27) --
				(102.31, 57.29) --
				(126.74, 65.34) --
				(151.16, 73.63) --
				(175.59, 82.65) --
				(200.02, 92.80) --
				(224.44,106.52);
			\definecolor{drawColor}{RGB}{37,122,164}
			
			\path[draw=drawColor,line width= 0.6pt,dash pattern=on 4pt off 2pt ,line join=round] ( 53.46, 34.30) --
				( 77.88, 44.15) --
				(102.31, 57.22) --
				(126.74, 76.21) --
				(151.16, 99.54);
			\definecolor{fillColor}{RGB}{37,122,164}
			
			\path[fill=fillColor] ( 75.92, 42.18) --
				( 79.85, 42.18) --
				( 79.85, 46.11) --
				( 75.92, 46.11) --
				cycle;
			
			\path[fill=fillColor] (149.20, 97.57) --
				(153.13, 97.57) --
				(153.13,101.50) --
				(149.20,101.50) --
				cycle;
			
			\path[fill=fillColor] ( 51.50, 32.33) --
				( 55.42, 32.33) --
				( 55.42, 36.26) --
				( 51.50, 36.26) --
				cycle;
			
			\path[fill=fillColor] (100.35, 55.25) --
				(104.27, 55.25) --
				(104.27, 59.18) --
				(100.35, 59.18) --
				cycle;
			
			\path[fill=fillColor] (124.77, 74.25) --
				(128.70, 74.25) --
				(128.70, 78.17) --
				(124.77, 78.17) --
				cycle;
			\definecolor{fillColor}{RGB}{247,192,26}
			
			\path[fill=fillColor] ( 77.88, 44.89) circle (  1.96);
			
			\path[fill=fillColor] (151.16, 99.58) circle (  1.96);
			
			\path[fill=fillColor] ( 53.46, 35.57) circle (  1.96);
			
			\path[fill=fillColor] (102.31, 57.55) circle (  1.96);
			
			\path[fill=fillColor] (126.74, 76.22) circle (  1.96);
			\definecolor{fillColor}{RGB}{78,155,133}
			
			\path[fill=fillColor] (200.02, 95.85) --
				(202.66, 91.27) --
				(197.37, 91.27) --
				cycle;
			
			\path[fill=fillColor] ( 77.88, 53.33) --
				( 80.53, 48.75) --
				( 75.24, 48.75) --
				cycle;
			
			\path[fill=fillColor] (175.59, 85.70) --
				(178.23, 81.12) --
				(172.95, 81.12) --
				cycle;
			
			\path[fill=fillColor] (151.16, 76.68) --
				(153.81, 72.10) --
				(148.52, 72.10) --
				cycle;
			
			\path[fill=fillColor] ( 53.46, 49.21) --
				( 56.10, 44.64) --
				( 50.82, 44.64) --
				cycle;
			
			\path[fill=fillColor] (102.31, 60.34) --
				(104.95, 55.77) --
				( 99.67, 55.77) --
				cycle;
			
			\path[fill=fillColor] (224.44,109.57) --
				(227.08,105.00) --
				(221.80,105.00) --
				cycle;
			
			\path[fill=fillColor] (126.74, 68.39) --
				(129.38, 63.81) --
				(124.09, 63.81) --
				cycle;
			\end{scope}
			\begin{scope}
			\path[clip] (  0.00,  0.00) rectangle (238.49,115.63);
			\definecolor{drawColor}{RGB}{0,0,0}
			
			\path[draw=drawColor,line width= 0.6pt,line join=round] ( 44.91, 30.69) --
				( 44.91,110.13);
			
			\path[draw=drawColor,line width= 0.6pt,line join=round] ( 46.33,107.67) --
				( 44.91,110.13) --
				( 43.49,107.67);
			\end{scope}
			\begin{scope}
			\path[clip] (  0.00,  0.00) rectangle (238.49,115.63);
			\definecolor{drawColor}{gray}{0.30}
			
			\node[text=drawColor,anchor=base east,inner sep=0pt, outer sep=0pt, scale=  0.88] at ( 39.96, 46.12) {1};
			
			\node[text=drawColor,anchor=base east,inner sep=0pt, outer sep=0pt, scale=  0.88] at ( 39.96, 70.96) {100};
			
			\node[text=drawColor,anchor=base east,inner sep=0pt, outer sep=0pt, scale=  0.88] at ( 39.96, 95.81) {10000};
			\end{scope}
			\begin{scope}
			\path[clip] (  0.00,  0.00) rectangle (238.49,115.63);
			\definecolor{drawColor}{gray}{0.20}
			
			\path[draw=drawColor,line width= 0.6pt,line join=round] ( 42.16, 49.15) --
				( 44.91, 49.15);
			
			\path[draw=drawColor,line width= 0.6pt,line join=round] ( 42.16, 73.99) --
				( 44.91, 73.99);
			
			\path[draw=drawColor,line width= 0.6pt,line join=round] ( 42.16, 98.84) --
				( 44.91, 98.84);
			\end{scope}
			\begin{scope}
			\path[clip] (  0.00,  0.00) rectangle (238.49,115.63);
			\definecolor{drawColor}{RGB}{0,0,0}
			
			\path[draw=drawColor,line width= 0.6pt,line join=round] ( 44.91, 30.69) --
				(232.99, 30.69);
			
			\path[draw=drawColor,line width= 0.6pt,line join=round] (230.53, 29.26) --
				(232.99, 30.69) --
				(230.53, 32.11);
			\end{scope}
			\begin{scope}
			\path[clip] (  0.00,  0.00) rectangle (238.49,115.63);
			\definecolor{drawColor}{gray}{0.20}
			
			\path[draw=drawColor,line width= 0.6pt,line join=round] ( 77.88, 27.94) --
				( 77.88, 30.69);
			
			\path[draw=drawColor,line width= 0.6pt,line join=round] (126.74, 27.94) --
				(126.74, 30.69);
			
			\path[draw=drawColor,line width= 0.6pt,line join=round] (175.59, 27.94) --
				(175.59, 30.69);
			
			\path[draw=drawColor,line width= 0.6pt,line join=round] (224.44, 27.94) --
				(224.44, 30.69);
			\end{scope}
			\begin{scope}
			\path[clip] (  0.00,  0.00) rectangle (238.49,115.63);
			\definecolor{drawColor}{gray}{0.30}
			
			\node[text=drawColor,anchor=base,inner sep=0pt, outer sep=0pt, scale=  0.88] at ( 77.88, 19.68) {4};
			
			\node[text=drawColor,anchor=base,inner sep=0pt, outer sep=0pt, scale=  0.88] at (126.74, 19.68) {8};
			
			\node[text=drawColor,anchor=base,inner sep=0pt, outer sep=0pt, scale=  0.88] at (175.59, 19.68) {12};
			
			\node[text=drawColor,anchor=base,inner sep=0pt, outer sep=0pt, scale=  0.88] at (224.44, 19.68) {16};
			\end{scope}
			\begin{scope}
			\path[clip] (  0.00,  0.00) rectangle (238.49,115.63);
			\definecolor{drawColor}{RGB}{0,0,0}
			
			\node[text=drawColor,anchor=base,inner sep=0pt, outer sep=0pt, scale=  1.10] at (138.95,  7.64) {$n$};
			\end{scope}
			\begin{scope}
			\path[clip] (  0.00,  0.00) rectangle (238.49,115.63);
			\definecolor{drawColor}{RGB}{0,0,0}
			
			\node[text=drawColor,rotate= 90.00,anchor=base,inner sep=0pt, outer sep=0pt, scale=  1.10] at ( 13.08, 70.41) {time (ms)};
			\end{scope}
			\begin{scope}
			\path[clip] (  0.00,  0.00) rectangle (238.49,115.63);
			
			\path[] ( 42.61, 63.09) rectangle ( 96.11,117.45);
			\end{scope}
			\begin{scope}
			\path[clip] (  0.00,  0.00) rectangle (238.49,115.63);
			\definecolor{drawColor}{RGB}{247,192,26}
			
			\path[draw=drawColor,line width= 0.6pt,line join=round] ( 49.55,104.72) -- ( 61.12,104.72);
			\end{scope}
			\begin{scope}
			\path[clip] (  0.00,  0.00) rectangle (238.49,115.63);
			\definecolor{fillColor}{RGB}{247,192,26}
			
			\path[fill=fillColor] ( 55.33,104.72) circle (  1.96);
			\end{scope}
			\begin{scope}
			\path[clip] (  0.00,  0.00) rectangle (238.49,115.63);
			\definecolor{drawColor}{RGB}{78,155,133}
			
			\path[draw=drawColor,line width= 0.6pt,dash pattern=on 2pt off 2pt ,line join=round] ( 49.55, 90.27) -- ( 61.12, 90.27);
			\end{scope}
			\begin{scope}
			\path[clip] (  0.00,  0.00) rectangle (238.49,115.63);
			\definecolor{fillColor}{RGB}{78,155,133}
			
			\path[fill=fillColor] ( 55.33, 93.32) --
				( 57.98, 88.74) --
				( 52.69, 88.74) --
				cycle;
			\end{scope}
			\begin{scope}
			\path[clip] (  0.00,  0.00) rectangle (238.49,115.63);
			\definecolor{drawColor}{RGB}{37,122,164}
			
			\path[draw=drawColor,line width= 0.6pt,dash pattern=on 4pt off 2pt ,line join=round] ( 49.55, 75.82) -- ( 61.12, 75.82);
			\end{scope}
			\begin{scope}
			\path[clip] (  0.00,  0.00) rectangle (238.49,115.63);
			\definecolor{fillColor}{RGB}{37,122,164}
			
			\path[fill=fillColor] ( 53.37, 73.85) --
				( 57.30, 73.85) --
				( 57.30, 77.78) --
				( 53.37, 77.78) --
				cycle;
			\end{scope}
			\begin{scope}
			\path[clip] (  0.00,  0.00) rectangle (238.49,115.63);
			\definecolor{drawColor}{RGB}{0,0,0}
			
			\node[text=drawColor,anchor=base west,inner sep=0pt, outer sep=0pt, scale=  0.80] at ( 68.06,101.97) {ACP};
			\end{scope}
			\begin{scope}
			\path[clip] (  0.00,  0.00) rectangle (238.49,115.63);
			\definecolor{drawColor}{RGB}{0,0,0}
			
			\node[text=drawColor,anchor=base west,inner sep=0pt, outer sep=0pt, scale=  0.80] at ( 68.06, 87.52) {DEFT};
			\end{scope}
			\begin{scope}
			\path[clip] (  0.00,  0.00) rectangle (238.49,115.63);
			\definecolor{drawColor}{RGB}{0,0,0}
			
			\node[text=drawColor,anchor=base west,inner sep=0pt, outer sep=0pt, scale=  0.80] at ( 68.06, 73.06) {Naive};
			\end{scope}
		\end{tikzpicture}
		\caption{Run times of the \enquote{naive} approach, the approach used in \ac{cpr}, and \ac{deft} for different numbers of arguments $n$ and a proportion $p = 0.0$ of identical potentials. The left plot shows results for non-exchangeable instances and the right plot presents results for exchangeable instances. Both plots use a logarithmic scale.}
		\label{fig:plot-asc}
	\end{figure*}
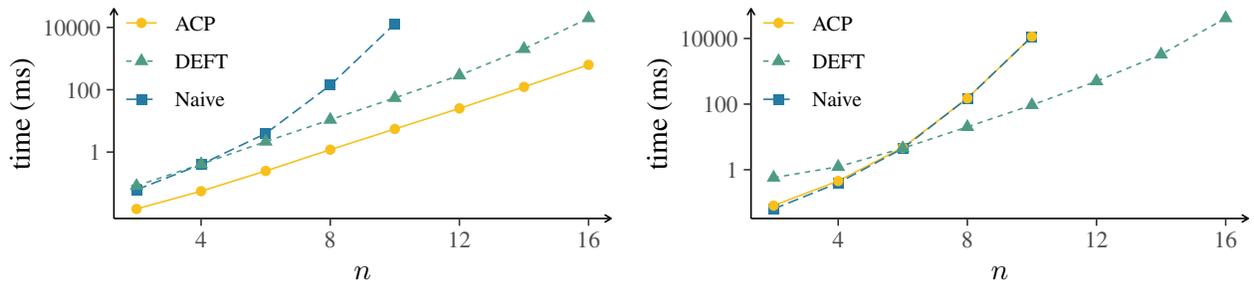

	In addition to the experimental results provided in \cref{sec:deft_experiments}, we give further plots for individual scenarios in this section.
	\Cref{fig:plot-asc} shows the run times for instances with a proportion of $p = 0.0$ identical potentials, i.e., each input factor maps each assignment of its arguments to a different potential value.
	The plot on the left shows the results for factors that are not exchangeable and the plot on the right depicts results for exchangeable factors.
	For both exchangeable and non-exchangeable factors, the naive approach is able to handle instances with up to $n = 10$ arguments and fails to solve larger instances.
	While \ac{cpr} exhibits a similar behaviour as the naive approach for exchangeable factors, \ac{cpr} is able to solve all non-exchangeable instances.
	This is expected as all non-exchangeable instances are generated such that the buckets of the factors under consideration are not mapped to the same multiset of potential values and thus, \ac{cpr} is able to return before iterating over any permutations.
	The left plot also indicates that \ac{deft} induces some overhead when iterating over the buckets, which is in line with our expectation as \ac{deft} performs additional pre-processing and computes intersections of possible argument rearrangements while \ac{cpr} does not.
	At the same time, \ac{deft} is able to solve all instances independent of whether the factors are exchangeable or not whereas \ac{cpr} solves larger instances only for non-exchangeable factors within the specified timeout.

	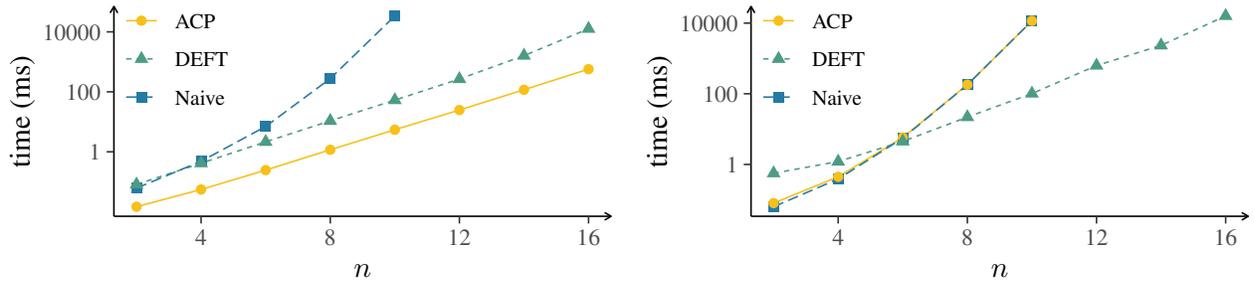
\begin{figure*}[t]
		\centering
		\begin{tikzpicture}[x=1pt,y=1pt]
			\definecolor{fillColor}{RGB}{255,255,255}
			\path[use as bounding box,fill=fillColor,fill opacity=0.00] (0,0) rectangle (238.49,115.63);
			\begin{scope}
			\path[clip] (  0.00,  0.00) rectangle (238.49,115.63);
			\definecolor{drawColor}{RGB}{255,255,255}
			\definecolor{fillColor}{RGB}{255,255,255}
			
			\path[draw=drawColor,line width= 0.6pt,line join=round,line cap=round,fill=fillColor] (  0.00,  0.00) rectangle (238.49,115.63);
			\end{scope}
			\begin{scope}
			\path[clip] ( 44.91, 30.69) rectangle (232.99,110.13);
			\definecolor{fillColor}{RGB}{255,255,255}
			
			\path[fill=fillColor] ( 44.91, 30.69) rectangle (232.99,110.13);
			\definecolor{drawColor}{RGB}{247,192,26}
			
			\path[draw=drawColor,line width= 0.6pt,line join=round] ( 53.46, 34.30) --
				( 77.88, 40.82) --
				(102.31, 48.16) --
				(126.74, 55.87) --
				(151.16, 63.45) --
				(175.59, 70.91) --
				(200.02, 78.58) --
				(224.44, 86.38);
			\definecolor{drawColor}{RGB}{78,155,133}
			
			\path[draw=drawColor,line width= 0.6pt,dash pattern=on 2pt off 2pt ,line join=round] ( 53.46, 42.73) --
				( 77.88, 50.77) --
				(102.31, 58.86) --
				(126.74, 66.81) --
				(151.16, 74.66) --
				(175.59, 82.64) --
				(200.02, 91.49) --
				(224.44,101.67);
			\definecolor{drawColor}{RGB}{37,122,164}
			
			\path[draw=drawColor,line width= 0.6pt,dash pattern=on 4pt off 2pt ,line join=round] ( 53.46, 41.34) --
				( 77.88, 51.50) --
				(102.31, 64.68) --
				(126.74, 82.81) --
				(151.16,106.52);
			\definecolor{fillColor}{RGB}{37,122,164}
			
			\path[fill=fillColor] ( 75.92, 49.54) --
				( 79.85, 49.54) --
				( 79.85, 53.47) --
				( 75.92, 53.47) --
				cycle;
			
			\path[fill=fillColor] (149.20,104.56) --
				(153.13,104.56) --
				(153.13,108.48) --
				(149.20,108.48) --
				cycle;
			
			\path[fill=fillColor] ( 51.50, 39.37) --
				( 55.42, 39.37) --
				( 55.42, 43.30) --
				( 51.50, 43.30) --
				cycle;
			
			\path[fill=fillColor] (100.35, 62.71) --
				(104.27, 62.71) --
				(104.27, 66.64) --
				(100.35, 66.64) --
				cycle;
			
			\path[fill=fillColor] (124.77, 80.85) --
				(128.70, 80.85) --
				(128.70, 84.77) --
				(124.77, 84.77) --
				cycle;
			\definecolor{fillColor}{RGB}{247,192,26}
			
			\path[fill=fillColor] (200.02, 78.58) circle (  1.96);
			
			\path[fill=fillColor] ( 77.88, 40.82) circle (  1.96);
			
			\path[fill=fillColor] (175.59, 70.91) circle (  1.96);
			
			\path[fill=fillColor] (151.16, 63.45) circle (  1.96);
			
			\path[fill=fillColor] ( 53.46, 34.30) circle (  1.96);
			
			\path[fill=fillColor] (102.31, 48.16) circle (  1.96);
			
			\path[fill=fillColor] (224.44, 86.38) circle (  1.96);
			
			\path[fill=fillColor] (126.74, 55.87) circle (  1.96);
			\definecolor{fillColor}{RGB}{78,155,133}
			
			\path[fill=fillColor] (200.02, 94.55) --
				(202.66, 89.97) --
				(197.37, 89.97) --
				cycle;
			
			\path[fill=fillColor] ( 77.88, 53.82) --
				( 80.53, 49.24) --
				( 75.24, 49.24) --
				cycle;
			
			\path[fill=fillColor] (175.59, 85.69) --
				(178.23, 81.11) --
				(172.95, 81.11) --
				cycle;
			
			\path[fill=fillColor] (151.16, 77.71) --
				(153.81, 73.14) --
				(148.52, 73.14) --
				cycle;
			
			\path[fill=fillColor] ( 53.46, 45.78) --
				( 56.10, 41.21) --
				( 50.82, 41.21) --
				cycle;
			
			\path[fill=fillColor] (102.31, 61.91) --
				(104.95, 57.33) --
				( 99.67, 57.33) --
				cycle;
			
			\path[fill=fillColor] (224.44,104.72) --
				(227.08,100.14) --
				(221.80,100.14) --
				cycle;
			
			\path[fill=fillColor] (126.74, 69.86) --
				(129.38, 65.28) --
				(124.09, 65.28) --
				cycle;
			\end{scope}
			\begin{scope}
			\path[clip] (  0.00,  0.00) rectangle (238.49,115.63);
			\definecolor{drawColor}{RGB}{0,0,0}
			
			\path[draw=drawColor,line width= 0.6pt,line join=round] ( 44.91, 30.69) --
				( 44.91,110.13);
			
			\path[draw=drawColor,line width= 0.6pt,line join=round] ( 46.33,107.67) --
				( 44.91,110.13) --
				( 43.49,107.67);
			\end{scope}
			\begin{scope}
			\path[clip] (  0.00,  0.00) rectangle (238.49,115.63);
			\definecolor{drawColor}{gray}{0.30}
			
			\node[text=drawColor,anchor=base east,inner sep=0pt, outer sep=0pt, scale=  0.88] at ( 39.96, 52.06) {1};
			
			\node[text=drawColor,anchor=base east,inner sep=0pt, outer sep=0pt, scale=  0.88] at ( 39.96, 74.77) {100};
			
			\node[text=drawColor,anchor=base east,inner sep=0pt, outer sep=0pt, scale=  0.88] at ( 39.96, 97.47) {10000};
			\end{scope}
			\begin{scope}
			\path[clip] (  0.00,  0.00) rectangle (238.49,115.63);
			\definecolor{drawColor}{gray}{0.20}
			
			\path[draw=drawColor,line width= 0.6pt,line join=round] ( 42.16, 55.09) --
				( 44.91, 55.09);
			
			\path[draw=drawColor,line width= 0.6pt,line join=round] ( 42.16, 77.80) --
				( 44.91, 77.80);
			
			\path[draw=drawColor,line width= 0.6pt,line join=round] ( 42.16,100.50) --
				( 44.91,100.50);
			\end{scope}
			\begin{scope}
			\path[clip] (  0.00,  0.00) rectangle (238.49,115.63);
			\definecolor{drawColor}{RGB}{0,0,0}
			
			\path[draw=drawColor,line width= 0.6pt,line join=round] ( 44.91, 30.69) --
				(232.99, 30.69);
			
			\path[draw=drawColor,line width= 0.6pt,line join=round] (230.53, 29.26) --
				(232.99, 30.69) --
				(230.53, 32.11);
			\end{scope}
			\begin{scope}
			\path[clip] (  0.00,  0.00) rectangle (238.49,115.63);
			\definecolor{drawColor}{gray}{0.20}
			
			\path[draw=drawColor,line width= 0.6pt,line join=round] ( 77.88, 27.94) --
				( 77.88, 30.69);
			
			\path[draw=drawColor,line width= 0.6pt,line join=round] (126.74, 27.94) --
				(126.74, 30.69);
			
			\path[draw=drawColor,line width= 0.6pt,line join=round] (175.59, 27.94) --
				(175.59, 30.69);
			
			\path[draw=drawColor,line width= 0.6pt,line join=round] (224.44, 27.94) --
				(224.44, 30.69);
			\end{scope}
			\begin{scope}
			\path[clip] (  0.00,  0.00) rectangle (238.49,115.63);
			\definecolor{drawColor}{gray}{0.30}
			
			\node[text=drawColor,anchor=base,inner sep=0pt, outer sep=0pt, scale=  0.88] at ( 77.88, 19.68) {4};
			
			\node[text=drawColor,anchor=base,inner sep=0pt, outer sep=0pt, scale=  0.88] at (126.74, 19.68) {8};
			
			\node[text=drawColor,anchor=base,inner sep=0pt, outer sep=0pt, scale=  0.88] at (175.59, 19.68) {12};
			
			\node[text=drawColor,anchor=base,inner sep=0pt, outer sep=0pt, scale=  0.88] at (224.44, 19.68) {16};
			\end{scope}
			\begin{scope}
			\path[clip] (  0.00,  0.00) rectangle (238.49,115.63);
			\definecolor{drawColor}{RGB}{0,0,0}
			
			\node[text=drawColor,anchor=base,inner sep=0pt, outer sep=0pt, scale=  1.10] at (138.95,  7.64) {$n$};
			\end{scope}
			\begin{scope}
			\path[clip] (  0.00,  0.00) rectangle (238.49,115.63);
			\definecolor{drawColor}{RGB}{0,0,0}
			
			\node[text=drawColor,rotate= 90.00,anchor=base,inner sep=0pt, outer sep=0pt, scale=  1.10] at ( 13.08, 70.41) {time (ms)};
			\end{scope}
			\begin{scope}
			\path[clip] (  0.00,  0.00) rectangle (238.49,115.63);
			
			\path[] ( 42.61, 63.09) rectangle ( 96.11,117.45);
			\end{scope}
			\begin{scope}
			\path[clip] (  0.00,  0.00) rectangle (238.49,115.63);
			\definecolor{drawColor}{RGB}{247,192,26}
			
			\path[draw=drawColor,line width= 0.6pt,line join=round] ( 49.55,104.72) -- ( 61.12,104.72);
			\end{scope}
			\begin{scope}
			\path[clip] (  0.00,  0.00) rectangle (238.49,115.63);
			\definecolor{fillColor}{RGB}{247,192,26}
			
			\path[fill=fillColor] ( 55.33,104.72) circle (  1.96);
			\end{scope}
			\begin{scope}
			\path[clip] (  0.00,  0.00) rectangle (238.49,115.63);
			\definecolor{drawColor}{RGB}{78,155,133}
			
			\path[draw=drawColor,line width= 0.6pt,dash pattern=on 2pt off 2pt ,line join=round] ( 49.55, 90.27) -- ( 61.12, 90.27);
			\end{scope}
			\begin{scope}
			\path[clip] (  0.00,  0.00) rectangle (238.49,115.63);
			\definecolor{fillColor}{RGB}{78,155,133}
			
			\path[fill=fillColor] ( 55.33, 93.32) --
				( 57.98, 88.74) --
				( 52.69, 88.74) --
				cycle;
			\end{scope}
			\begin{scope}
			\path[clip] (  0.00,  0.00) rectangle (238.49,115.63);
			\definecolor{drawColor}{RGB}{37,122,164}
			
			\path[draw=drawColor,line width= 0.6pt,dash pattern=on 4pt off 2pt ,line join=round] ( 49.55, 75.82) -- ( 61.12, 75.82);
			\end{scope}
			\begin{scope}
			\path[clip] (  0.00,  0.00) rectangle (238.49,115.63);
			\definecolor{fillColor}{RGB}{37,122,164}
			
			\path[fill=fillColor] ( 53.37, 73.85) --
				( 57.30, 73.85) --
				( 57.30, 77.78) --
				( 53.37, 77.78) --
				cycle;
			\end{scope}
			\begin{scope}
			\path[clip] (  0.00,  0.00) rectangle (238.49,115.63);
			\definecolor{drawColor}{RGB}{0,0,0}
			
			\node[text=drawColor,anchor=base west,inner sep=0pt, outer sep=0pt, scale=  0.80] at ( 68.06,101.97) {ACP};
			\end{scope}
			\begin{scope}
			\path[clip] (  0.00,  0.00) rectangle (238.49,115.63);
			\definecolor{drawColor}{RGB}{0,0,0}
			
			\node[text=drawColor,anchor=base west,inner sep=0pt, outer sep=0pt, scale=  0.80] at ( 68.06, 87.52) {DEFT};
			\end{scope}
			\begin{scope}
			\path[clip] (  0.00,  0.00) rectangle (238.49,115.63);
			\definecolor{drawColor}{RGB}{0,0,0}
			
			\node[text=drawColor,anchor=base west,inner sep=0pt, outer sep=0pt, scale=  0.80] at ( 68.06, 73.06) {Naive};
			\end{scope}
		\end{tikzpicture}
		\begin{tikzpicture}[x=1pt,y=1pt]
			\definecolor{fillColor}{RGB}{255,255,255}
			\path[use as bounding box,fill=fillColor,fill opacity=0.00] (0,0) rectangle (238.49,115.63);
			\begin{scope}
			\path[clip] (  0.00,  0.00) rectangle (238.49,115.63);
			\definecolor{drawColor}{RGB}{255,255,255}
			\definecolor{fillColor}{RGB}{255,255,255}
			
			\path[draw=drawColor,line width= 0.6pt,line join=round,line cap=round,fill=fillColor] (  0.00,  0.00) rectangle (238.49,115.63);
			\end{scope}
			\begin{scope}
			\path[clip] ( 44.91, 30.69) rectangle (232.99,110.13);
			\definecolor{fillColor}{RGB}{255,255,255}
			
			\path[fill=fillColor] ( 44.91, 30.69) rectangle (232.99,110.13);
			\definecolor{drawColor}{RGB}{247,192,26}
			
			\path[draw=drawColor,line width= 0.6pt,line join=round] ( 53.46, 35.68) --
				( 77.88, 45.64) --
				(102.31, 60.52) --
				(126.74, 80.48) --
				(151.16,104.66);
			\definecolor{drawColor}{RGB}{78,155,133}
			
			\path[draw=drawColor,line width= 0.6pt,dash pattern=on 2pt off 2pt ,line join=round] ( 53.46, 46.98) --
				( 77.88, 51.40) --
				(102.31, 59.03) --
				(126.74, 68.24) --
				(151.16, 77.14) --
				(175.59, 87.73) --
				(200.02, 95.38) --
				(224.44,106.52);
			\definecolor{drawColor}{RGB}{37,122,164}
			
			\path[draw=drawColor,line width= 0.6pt,dash pattern=on 4pt off 2pt ,line join=round] ( 53.46, 34.30) --
				( 77.88, 44.82) --
				(102.31, 60.25) --
				(126.74, 80.47) --
				(151.16,104.65);
			\definecolor{fillColor}{RGB}{37,122,164}
			
			\path[fill=fillColor] ( 75.92, 42.85) --
				( 79.85, 42.85) --
				( 79.85, 46.78) --
				( 75.92, 46.78) --
				cycle;
			
			\path[fill=fillColor] (149.20,102.69) --
				(153.13,102.69) --
				(153.13,106.61) --
				(149.20,106.61) --
				cycle;
			
			\path[fill=fillColor] ( 51.50, 32.33) --
				( 55.42, 32.33) --
				( 55.42, 36.26) --
				( 51.50, 36.26) --
				cycle;
			
			\path[fill=fillColor] (100.35, 58.29) --
				(104.27, 58.29) --
				(104.27, 62.21) --
				(100.35, 62.21) --
				cycle;
			
			\path[fill=fillColor] (124.77, 78.51) --
				(128.70, 78.51) --
				(128.70, 82.43) --
				(124.77, 82.43) --
				cycle;
			\definecolor{fillColor}{RGB}{247,192,26}
			
			\path[fill=fillColor] ( 77.88, 45.64) circle (  1.96);
			
			\path[fill=fillColor] (151.16,104.66) circle (  1.96);
			
			\path[fill=fillColor] ( 53.46, 35.68) circle (  1.96);
			
			\path[fill=fillColor] (102.31, 60.52) circle (  1.96);
			
			\path[fill=fillColor] (126.74, 80.48) circle (  1.96);
			\definecolor{fillColor}{RGB}{78,155,133}
			
			\path[fill=fillColor] (200.02, 98.43) --
				(202.66, 93.85) --
				(197.37, 93.85) --
				cycle;
			
			\path[fill=fillColor] ( 77.88, 54.45) --
				( 80.53, 49.87) --
				( 75.24, 49.87) --
				cycle;
			
			\path[fill=fillColor] (175.59, 90.79) --
				(178.23, 86.21) --
				(172.95, 86.21) --
				cycle;
			
			\path[fill=fillColor] (151.16, 80.19) --
				(153.81, 75.62) --
				(148.52, 75.62) --
				cycle;
			
			\path[fill=fillColor] ( 53.46, 50.03) --
				( 56.10, 45.45) --
				( 50.82, 45.45) --
				cycle;
			
			\path[fill=fillColor] (102.31, 62.08) --
				(104.95, 57.50) --
				( 99.67, 57.50) --
				cycle;
			
			\path[fill=fillColor] (224.44,109.57) --
				(227.08,105.00) --
				(221.80,105.00) --
				cycle;
			
			\path[fill=fillColor] (126.74, 71.29) --
				(129.38, 66.71) --
				(124.09, 66.71) --
				cycle;
			\end{scope}
			\begin{scope}
			\path[clip] (  0.00,  0.00) rectangle (238.49,115.63);
			\definecolor{drawColor}{RGB}{0,0,0}
			
			\path[draw=drawColor,line width= 0.6pt,line join=round] ( 44.91, 30.69) --
				( 44.91,110.13);
			
			\path[draw=drawColor,line width= 0.6pt,line join=round] ( 46.33,107.67) --
				( 44.91,110.13) --
				( 43.49,107.67);
			\end{scope}
			\begin{scope}
			\path[clip] (  0.00,  0.00) rectangle (238.49,115.63);
			\definecolor{drawColor}{gray}{0.30}
			
			\node[text=drawColor,anchor=base east,inner sep=0pt, outer sep=0pt, scale=  0.88] at ( 39.96, 47.28) {1};
			
			\node[text=drawColor,anchor=base east,inner sep=0pt, outer sep=0pt, scale=  0.88] at ( 39.96, 74.03) {100};
			
			\node[text=drawColor,anchor=base east,inner sep=0pt, outer sep=0pt, scale=  0.88] at ( 39.96,100.78) {10000};
			\end{scope}
			\begin{scope}
			\path[clip] (  0.00,  0.00) rectangle (238.49,115.63);
			\definecolor{drawColor}{gray}{0.20}
			
			\path[draw=drawColor,line width= 0.6pt,line join=round] ( 42.16, 50.31) --
				( 44.91, 50.31);
			
			\path[draw=drawColor,line width= 0.6pt,line join=round] ( 42.16, 77.06) --
				( 44.91, 77.06);
			
			\path[draw=drawColor,line width= 0.6pt,line join=round] ( 42.16,103.81) --
				( 44.91,103.81);
			\end{scope}
			\begin{scope}
			\path[clip] (  0.00,  0.00) rectangle (238.49,115.63);
			\definecolor{drawColor}{RGB}{0,0,0}
			
			\path[draw=drawColor,line width= 0.6pt,line join=round] ( 44.91, 30.69) --
				(232.99, 30.69);
			
			\path[draw=drawColor,line width= 0.6pt,line join=round] (230.53, 29.26) --
				(232.99, 30.69) --
				(230.53, 32.11);
			\end{scope}
			\begin{scope}
			\path[clip] (  0.00,  0.00) rectangle (238.49,115.63);
			\definecolor{drawColor}{gray}{0.20}
			
			\path[draw=drawColor,line width= 0.6pt,line join=round] ( 77.88, 27.94) --
				( 77.88, 30.69);
			
			\path[draw=drawColor,line width= 0.6pt,line join=round] (126.74, 27.94) --
				(126.74, 30.69);
			
			\path[draw=drawColor,line width= 0.6pt,line join=round] (175.59, 27.94) --
				(175.59, 30.69);
			
			\path[draw=drawColor,line width= 0.6pt,line join=round] (224.44, 27.94) --
				(224.44, 30.69);
			\end{scope}
			\begin{scope}
			\path[clip] (  0.00,  0.00) rectangle (238.49,115.63);
			\definecolor{drawColor}{gray}{0.30}
			
			\node[text=drawColor,anchor=base,inner sep=0pt, outer sep=0pt, scale=  0.88] at ( 77.88, 19.68) {4};
			
			\node[text=drawColor,anchor=base,inner sep=0pt, outer sep=0pt, scale=  0.88] at (126.74, 19.68) {8};
			
			\node[text=drawColor,anchor=base,inner sep=0pt, outer sep=0pt, scale=  0.88] at (175.59, 19.68) {12};
			
			\node[text=drawColor,anchor=base,inner sep=0pt, outer sep=0pt, scale=  0.88] at (224.44, 19.68) {16};
			\end{scope}
			\begin{scope}
			\path[clip] (  0.00,  0.00) rectangle (238.49,115.63);
			\definecolor{drawColor}{RGB}{0,0,0}
			
			\node[text=drawColor,anchor=base,inner sep=0pt, outer sep=0pt, scale=  1.10] at (138.95,  7.64) {$n$};
			\end{scope}
			\begin{scope}
			\path[clip] (  0.00,  0.00) rectangle (238.49,115.63);
			\definecolor{drawColor}{RGB}{0,0,0}
			
			\node[text=drawColor,rotate= 90.00,anchor=base,inner sep=0pt, outer sep=0pt, scale=  1.10] at ( 13.08, 70.41) {time (ms)};
			\end{scope}
			\begin{scope}
			\path[clip] (  0.00,  0.00) rectangle (238.49,115.63);
			
			\path[] ( 42.61, 63.09) rectangle ( 96.11,117.45);
			\end{scope}
			\begin{scope}
			\path[clip] (  0.00,  0.00) rectangle (238.49,115.63);
			\definecolor{drawColor}{RGB}{247,192,26}
			
			\path[draw=drawColor,line width= 0.6pt,line join=round] ( 49.55,104.72) -- ( 61.12,104.72);
			\end{scope}
			\begin{scope}
			\path[clip] (  0.00,  0.00) rectangle (238.49,115.63);
			\definecolor{fillColor}{RGB}{247,192,26}
			
			\path[fill=fillColor] ( 55.33,104.72) circle (  1.96);
			\end{scope}
			\begin{scope}
			\path[clip] (  0.00,  0.00) rectangle (238.49,115.63);
			\definecolor{drawColor}{RGB}{78,155,133}
			
			\path[draw=drawColor,line width= 0.6pt,dash pattern=on 2pt off 2pt ,line join=round] ( 49.55, 90.27) -- ( 61.12, 90.27);
			\end{scope}
			\begin{scope}
			\path[clip] (  0.00,  0.00) rectangle (238.49,115.63);
			\definecolor{fillColor}{RGB}{78,155,133}
			
			\path[fill=fillColor] ( 55.33, 93.32) --
				( 57.98, 88.74) --
				( 52.69, 88.74) --
				cycle;
			\end{scope}
			\begin{scope}
			\path[clip] (  0.00,  0.00) rectangle (238.49,115.63);
			\definecolor{drawColor}{RGB}{37,122,164}
			
			\path[draw=drawColor,line width= 0.6pt,dash pattern=on 4pt off 2pt ,line join=round] ( 49.55, 75.82) -- ( 61.12, 75.82);
			\end{scope}
			\begin{scope}
			\path[clip] (  0.00,  0.00) rectangle (238.49,115.63);
			\definecolor{fillColor}{RGB}{37,122,164}
			
			\path[fill=fillColor] ( 53.37, 73.85) --
				( 57.30, 73.85) --
				( 57.30, 77.78) --
				( 53.37, 77.78) --
				cycle;
			\end{scope}
			\begin{scope}
			\path[clip] (  0.00,  0.00) rectangle (238.49,115.63);
			\definecolor{drawColor}{RGB}{0,0,0}
			
			\node[text=drawColor,anchor=base west,inner sep=0pt, outer sep=0pt, scale=  0.80] at ( 68.06,101.97) {ACP};
			\end{scope}
			\begin{scope}
			\path[clip] (  0.00,  0.00) rectangle (238.49,115.63);
			\definecolor{drawColor}{RGB}{0,0,0}
			
			\node[text=drawColor,anchor=base west,inner sep=0pt, outer sep=0pt, scale=  0.80] at ( 68.06, 87.52) {DEFT};
			\end{scope}
			\begin{scope}
			\path[clip] (  0.00,  0.00) rectangle (238.49,115.63);
			\definecolor{drawColor}{RGB}{0,0,0}
			
			\node[text=drawColor,anchor=base west,inner sep=0pt, outer sep=0pt, scale=  0.80] at ( 68.06, 73.06) {Naive};
			\end{scope}
		\end{tikzpicture}
		\caption{Run times of the \enquote{naive} approach, the approach used in \ac{cpr}, and \ac{deft} for different numbers of arguments $n$ and different proportions $p \in \{0.1, 0.2, 0.5, 0.8, 0.9\}$ of identical potentials (averaged). The left plot shows results for non-exchangeable instances and the right plot presents results for exchangeable instances. Both plots use a logarithmic scale.}
		\label{fig:plot-mixed}
	\end{figure*}

	\Cref{fig:plot-mixed} contains the average run times for instances with a proportion of $p \in \{0.1, 0.2, 0.5, 0.8, 0.9\}$ identical potentials.
	More specifically, every input factor maps each assignment of its arguments to the same potential value with a probability of $p$ (i.e., out of all rows in the table, there is roughly a proportion of $p$ rows that are mapped to the same potential value).
	We can see that both plots depicted in \cref{fig:plot-mixed} exhibit the same patterns as the plots in \cref{fig:plot-asc}.

	\begin{figure*}[t]
		\centering
		\begin{tikzpicture}[x=1pt,y=1pt]
			\definecolor{fillColor}{RGB}{255,255,255}
			\path[use as bounding box,fill=fillColor,fill opacity=0.00] (0,0) rectangle (238.49,115.63);
			\begin{scope}
			\path[clip] (  0.00,  0.00) rectangle (238.49,115.63);
			\definecolor{drawColor}{RGB}{255,255,255}
			\definecolor{fillColor}{RGB}{255,255,255}
			
			\path[draw=drawColor,line width= 0.6pt,line join=round,line cap=round,fill=fillColor] (  0.00,  0.00) rectangle (238.49,115.63);
			\end{scope}
			\begin{scope}
			\path[clip] ( 44.91, 30.69) rectangle (232.99,110.13);
			\definecolor{fillColor}{RGB}{255,255,255}
			
			\path[fill=fillColor] ( 44.91, 30.69) rectangle (232.99,110.13);
			\definecolor{drawColor}{RGB}{247,192,26}
			
			\path[draw=drawColor,line width= 0.6pt,line join=round] ( 53.46, 34.30) --
				( 77.88, 41.20) --
				(102.31, 49.01) --
				(126.74, 57.18) --
				(151.16, 65.24) --
				(175.59, 73.29) --
				(200.02, 81.45) --
				(224.44, 89.52);
			\definecolor{drawColor}{RGB}{78,155,133}
			
			\path[draw=drawColor,line width= 0.6pt,dash pattern=on 2pt off 2pt ,line join=round] ( 53.46, 43.34) --
				( 77.88, 51.75) --
				(102.31, 60.45) --
				(126.74, 68.97) --
				(151.16, 77.29) --
				(175.59, 85.47) --
				(200.02, 93.81) --
				(224.44,102.17);
			\definecolor{drawColor}{RGB}{37,122,164}
			
			\path[draw=drawColor,line width= 0.6pt,dash pattern=on 4pt off 2pt ,line join=round] ( 53.46, 41.81) --
				( 77.88, 51.69) --
				(102.31, 63.71) --
				(126.74, 82.50) --
				(151.16,106.52);
			\definecolor{fillColor}{RGB}{37,122,164}
			
			\path[fill=fillColor] ( 75.92, 49.73) --
				( 79.85, 49.73) --
				( 79.85, 53.65) --
				( 75.92, 53.65) --
				cycle;
			
			\path[fill=fillColor] (149.20,104.56) --
				(153.13,104.56) --
				(153.13,108.48) --
				(149.20,108.48) --
				cycle;
			
			\path[fill=fillColor] ( 51.50, 39.85) --
				( 55.42, 39.85) --
				( 55.42, 43.77) --
				( 51.50, 43.77) --
				cycle;
			
			\path[fill=fillColor] (100.35, 61.75) --
				(104.27, 61.75) --
				(104.27, 65.68) --
				(100.35, 65.68) --
				cycle;
			
			\path[fill=fillColor] (124.77, 80.54) --
				(128.70, 80.54) --
				(128.70, 84.46) --
				(124.77, 84.46) --
				cycle;
			\definecolor{fillColor}{RGB}{247,192,26}
			
			\path[fill=fillColor] (200.02, 81.45) circle (  1.96);
			
			\path[fill=fillColor] ( 77.88, 41.20) circle (  1.96);
			
			\path[fill=fillColor] (175.59, 73.29) circle (  1.96);
			
			\path[fill=fillColor] (151.16, 65.24) circle (  1.96);
			
			\path[fill=fillColor] ( 53.46, 34.30) circle (  1.96);
			
			\path[fill=fillColor] (102.31, 49.01) circle (  1.96);
			
			\path[fill=fillColor] (224.44, 89.52) circle (  1.96);
			
			\path[fill=fillColor] (126.74, 57.18) circle (  1.96);
			\definecolor{fillColor}{RGB}{78,155,133}
			
			\path[fill=fillColor] (200.02, 96.86) --
				(202.66, 92.28) --
				(197.37, 92.28) --
				cycle;
			
			\path[fill=fillColor] ( 77.88, 54.80) --
				( 80.53, 50.22) --
				( 75.24, 50.22) --
				cycle;
			
			\path[fill=fillColor] (175.59, 88.53) --
				(178.23, 83.95) --
				(172.95, 83.95) --
				cycle;
			
			\path[fill=fillColor] (151.16, 80.34) --
				(153.81, 75.76) --
				(148.52, 75.76) --
				cycle;
			
			\path[fill=fillColor] ( 53.46, 46.39) --
				( 56.10, 41.81) --
				( 50.82, 41.81) --
				cycle;
			
			\path[fill=fillColor] (102.31, 63.50) --
				(104.95, 58.93) --
				( 99.67, 58.93) --
				cycle;
			
			\path[fill=fillColor] (224.44,105.22) --
				(227.08,100.64) --
				(221.80,100.64) --
				cycle;
			
			\path[fill=fillColor] (126.74, 72.02) --
				(129.38, 67.45) --
				(124.09, 67.45) --
				cycle;
			\end{scope}
			\begin{scope}
			\path[clip] (  0.00,  0.00) rectangle (238.49,115.63);
			\definecolor{drawColor}{RGB}{0,0,0}
			
			\path[draw=drawColor,line width= 0.6pt,line join=round] ( 44.91, 30.69) --
				( 44.91,110.13);
			
			\path[draw=drawColor,line width= 0.6pt,line join=round] ( 46.33,107.67) --
				( 44.91,110.13) --
				( 43.49,107.67);
			\end{scope}
			\begin{scope}
			\path[clip] (  0.00,  0.00) rectangle (238.49,115.63);
			\definecolor{drawColor}{gray}{0.30}
			
			\node[text=drawColor,anchor=base east,inner sep=0pt, outer sep=0pt, scale=  0.88] at ( 39.96, 53.46) {1};
			
			\node[text=drawColor,anchor=base east,inner sep=0pt, outer sep=0pt, scale=  0.88] at ( 39.96, 77.73) {100};
			
			\node[text=drawColor,anchor=base east,inner sep=0pt, outer sep=0pt, scale=  0.88] at ( 39.96,102.01) {10000};
			\end{scope}
			\begin{scope}
			\path[clip] (  0.00,  0.00) rectangle (238.49,115.63);
			\definecolor{drawColor}{gray}{0.20}
			
			\path[draw=drawColor,line width= 0.6pt,line join=round] ( 42.16, 56.49) --
				( 44.91, 56.49);
			
			\path[draw=drawColor,line width= 0.6pt,line join=round] ( 42.16, 80.76) --
				( 44.91, 80.76);
			
			\path[draw=drawColor,line width= 0.6pt,line join=round] ( 42.16,105.04) --
				( 44.91,105.04);
			\end{scope}
			\begin{scope}
			\path[clip] (  0.00,  0.00) rectangle (238.49,115.63);
			\definecolor{drawColor}{RGB}{0,0,0}
			
			\path[draw=drawColor,line width= 0.6pt,line join=round] ( 44.91, 30.69) --
				(232.99, 30.69);
			
			\path[draw=drawColor,line width= 0.6pt,line join=round] (230.53, 29.26) --
				(232.99, 30.69) --
				(230.53, 32.11);
			\end{scope}
			\begin{scope}
			\path[clip] (  0.00,  0.00) rectangle (238.49,115.63);
			\definecolor{drawColor}{gray}{0.20}
			
			\path[draw=drawColor,line width= 0.6pt,line join=round] ( 77.88, 27.94) --
				( 77.88, 30.69);
			
			\path[draw=drawColor,line width= 0.6pt,line join=round] (126.74, 27.94) --
				(126.74, 30.69);
			
			\path[draw=drawColor,line width= 0.6pt,line join=round] (175.59, 27.94) --
				(175.59, 30.69);
			
			\path[draw=drawColor,line width= 0.6pt,line join=round] (224.44, 27.94) --
				(224.44, 30.69);
			\end{scope}
			\begin{scope}
			\path[clip] (  0.00,  0.00) rectangle (238.49,115.63);
			\definecolor{drawColor}{gray}{0.30}
			
			\node[text=drawColor,anchor=base,inner sep=0pt, outer sep=0pt, scale=  0.88] at ( 77.88, 19.68) {4};
			
			\node[text=drawColor,anchor=base,inner sep=0pt, outer sep=0pt, scale=  0.88] at (126.74, 19.68) {8};
			
			\node[text=drawColor,anchor=base,inner sep=0pt, outer sep=0pt, scale=  0.88] at (175.59, 19.68) {12};
			
			\node[text=drawColor,anchor=base,inner sep=0pt, outer sep=0pt, scale=  0.88] at (224.44, 19.68) {16};
			\end{scope}
			\begin{scope}
			\path[clip] (  0.00,  0.00) rectangle (238.49,115.63);
			\definecolor{drawColor}{RGB}{0,0,0}
			
			\node[text=drawColor,anchor=base,inner sep=0pt, outer sep=0pt, scale=  1.10] at (138.95,  7.64) {$n$};
			\end{scope}
			\begin{scope}
			\path[clip] (  0.00,  0.00) rectangle (238.49,115.63);
			\definecolor{drawColor}{RGB}{0,0,0}
			
			\node[text=drawColor,rotate= 90.00,anchor=base,inner sep=0pt, outer sep=0pt, scale=  1.10] at ( 13.08, 70.41) {time (ms)};
			\end{scope}
			\begin{scope}
			\path[clip] (  0.00,  0.00) rectangle (238.49,115.63);
			
			\path[] ( 42.61, 63.09) rectangle ( 96.11,117.45);
			\end{scope}
			\begin{scope}
			\path[clip] (  0.00,  0.00) rectangle (238.49,115.63);
			\definecolor{drawColor}{RGB}{247,192,26}
			
			\path[draw=drawColor,line width= 0.6pt,line join=round] ( 49.55,104.72) -- ( 61.12,104.72);
			\end{scope}
			\begin{scope}
			\path[clip] (  0.00,  0.00) rectangle (238.49,115.63);
			\definecolor{fillColor}{RGB}{247,192,26}
			
			\path[fill=fillColor] ( 55.33,104.72) circle (  1.96);
			\end{scope}
			\begin{scope}
			\path[clip] (  0.00,  0.00) rectangle (238.49,115.63);
			\definecolor{drawColor}{RGB}{78,155,133}
			
			\path[draw=drawColor,line width= 0.6pt,dash pattern=on 2pt off 2pt ,line join=round] ( 49.55, 90.27) -- ( 61.12, 90.27);
			\end{scope}
			\begin{scope}
			\path[clip] (  0.00,  0.00) rectangle (238.49,115.63);
			\definecolor{fillColor}{RGB}{78,155,133}
			
			\path[fill=fillColor] ( 55.33, 93.32) --
				( 57.98, 88.74) --
				( 52.69, 88.74) --
				cycle;
			\end{scope}
			\begin{scope}
			\path[clip] (  0.00,  0.00) rectangle (238.49,115.63);
			\definecolor{drawColor}{RGB}{37,122,164}
			
			\path[draw=drawColor,line width= 0.6pt,dash pattern=on 4pt off 2pt ,line join=round] ( 49.55, 75.82) -- ( 61.12, 75.82);
			\end{scope}
			\begin{scope}
			\path[clip] (  0.00,  0.00) rectangle (238.49,115.63);
			\definecolor{fillColor}{RGB}{37,122,164}
			
			\path[fill=fillColor] ( 53.37, 73.85) --
				( 57.30, 73.85) --
				( 57.30, 77.78) --
				( 53.37, 77.78) --
				cycle;
			\end{scope}
			\begin{scope}
			\path[clip] (  0.00,  0.00) rectangle (238.49,115.63);
			\definecolor{drawColor}{RGB}{0,0,0}
			
			\node[text=drawColor,anchor=base west,inner sep=0pt, outer sep=0pt, scale=  0.80] at ( 68.06,101.97) {ACP};
			\end{scope}
			\begin{scope}
			\path[clip] (  0.00,  0.00) rectangle (238.49,115.63);
			\definecolor{drawColor}{RGB}{0,0,0}
			
			\node[text=drawColor,anchor=base west,inner sep=0pt, outer sep=0pt, scale=  0.80] at ( 68.06, 87.52) {DEFT};
			\end{scope}
			\begin{scope}
			\path[clip] (  0.00,  0.00) rectangle (238.49,115.63);
			\definecolor{drawColor}{RGB}{0,0,0}
			
			\node[text=drawColor,anchor=base west,inner sep=0pt, outer sep=0pt, scale=  0.80] at ( 68.06, 73.06) {Naive};
			\end{scope}
		\end{tikzpicture}
		\begin{tikzpicture}[x=1pt,y=1pt]
			\definecolor{fillColor}{RGB}{255,255,255}
			\path[use as bounding box,fill=fillColor,fill opacity=0.00] (0,0) rectangle (238.49,115.63);
			\begin{scope}
			\path[clip] (  0.00,  0.00) rectangle (238.49,115.63);
			\definecolor{drawColor}{RGB}{255,255,255}
			\definecolor{fillColor}{RGB}{255,255,255}
			
			\path[draw=drawColor,line width= 0.6pt,line join=round,line cap=round,fill=fillColor] (  0.00,  0.00) rectangle (238.49,115.63);
			\end{scope}
			\begin{scope}
			\path[clip] ( 44.91, 30.69) rectangle (232.99,110.13);
			\definecolor{fillColor}{RGB}{255,255,255}
			
			\path[fill=fillColor] ( 44.91, 30.69) rectangle (232.99,110.13);
			\definecolor{drawColor}{RGB}{247,192,26}
			
			\path[draw=drawColor,line width= 0.6pt,line join=round] ( 53.46, 35.72) --
				( 77.88, 45.84) --
				(102.31, 55.91) --
				(126.74, 65.70) --
				(151.16, 75.37) --
				(175.59, 84.80) --
				(200.02, 94.30) --
				(224.44,104.22);
			\definecolor{drawColor}{RGB}{78,155,133}
			
			\path[draw=drawColor,line width= 0.6pt,dash pattern=on 2pt off 2pt ,line join=round] ( 53.46, 47.39) --
				( 77.88, 51.56) --
				(102.31, 59.24) --
				(126.74, 68.39) --
				(151.16, 77.85) --
				(175.59, 87.31) --
				(200.02, 96.98) --
				(224.44,106.52);
			\definecolor{drawColor}{RGB}{37,122,164}
			
			\path[draw=drawColor,line width= 0.6pt,dash pattern=on 4pt off 2pt ,line join=round] ( 53.46, 34.30) --
				( 77.88, 44.92) --
				(102.31, 55.19) --
				(126.74, 65.02) --
				(151.16, 74.69) --
				(175.59, 84.21) --
				(200.02, 93.72) --
				(224.44,103.59);
			\definecolor{fillColor}{RGB}{37,122,164}
			
			\path[fill=fillColor] (198.05, 91.76) --
				(201.98, 91.76) --
				(201.98, 95.68) --
				(198.05, 95.68) --
				cycle;
			
			\path[fill=fillColor] ( 75.92, 42.96) --
				( 79.85, 42.96) --
				( 79.85, 46.88) --
				( 75.92, 46.88) --
				cycle;
			
			\path[fill=fillColor] (173.63, 82.25) --
				(177.55, 82.25) --
				(177.55, 86.17) --
				(173.63, 86.17) --
				cycle;
			
			\path[fill=fillColor] (149.20, 72.72) --
				(153.13, 72.72) --
				(153.13, 76.65) --
				(149.20, 76.65) --
				cycle;
			
			\path[fill=fillColor] ( 51.50, 32.33) --
				( 55.42, 32.33) --
				( 55.42, 36.26) --
				( 51.50, 36.26) --
				cycle;
			
			\path[fill=fillColor] (100.35, 53.23) --
				(104.27, 53.23) --
				(104.27, 57.16) --
				(100.35, 57.16) --
				cycle;
			
			\path[fill=fillColor] (222.48,101.62) --
				(226.40,101.62) --
				(226.40,105.55) --
				(222.48,105.55) --
				cycle;
			
			\path[fill=fillColor] (124.77, 63.06) --
				(128.70, 63.06) --
				(128.70, 66.98) --
				(124.77, 66.98) --
				cycle;
			\definecolor{fillColor}{RGB}{247,192,26}
			
			\path[fill=fillColor] (200.02, 94.30) circle (  1.96);
			
			\path[fill=fillColor] ( 77.88, 45.84) circle (  1.96);
			
			\path[fill=fillColor] (175.59, 84.80) circle (  1.96);
			
			\path[fill=fillColor] (151.16, 75.37) circle (  1.96);
			
			\path[fill=fillColor] ( 53.46, 35.72) circle (  1.96);
			
			\path[fill=fillColor] (102.31, 55.91) circle (  1.96);
			
			\path[fill=fillColor] (224.44,104.22) circle (  1.96);
			
			\path[fill=fillColor] (126.74, 65.70) circle (  1.96);
			\definecolor{fillColor}{RGB}{78,155,133}
			
			\path[fill=fillColor] (200.02,100.03) --
				(202.66, 95.45) --
				(197.37, 95.45) --
				cycle;
			
			\path[fill=fillColor] ( 77.88, 54.61) --
				( 80.53, 50.03) --
				( 75.24, 50.03) --
				cycle;
			
			\path[fill=fillColor] (175.59, 90.36) --
				(178.23, 85.78) --
				(172.95, 85.78) --
				cycle;
			
			\path[fill=fillColor] (151.16, 80.90) --
				(153.81, 76.33) --
				(148.52, 76.33) --
				cycle;
			
			\path[fill=fillColor] ( 53.46, 50.44) --
				( 56.10, 45.86) --
				( 50.82, 45.86) --
				cycle;
			
			\path[fill=fillColor] (102.31, 62.29) --
				(104.95, 57.71) --
				( 99.67, 57.71) --
				cycle;
			
			\path[fill=fillColor] (224.44,109.57) --
				(227.08,105.00) --
				(221.80,105.00) --
				cycle;
			
			\path[fill=fillColor] (126.74, 71.44) --
				(129.38, 66.87) --
				(124.09, 66.87) --
				cycle;
			\end{scope}
			\begin{scope}
			\path[clip] (  0.00,  0.00) rectangle (238.49,115.63);
			\definecolor{drawColor}{RGB}{0,0,0}
			
			\path[draw=drawColor,line width= 0.6pt,line join=round] ( 44.91, 30.69) --
				( 44.91,110.13);
			
			\path[draw=drawColor,line width= 0.6pt,line join=round] ( 46.33,107.67) --
				( 44.91,110.13) --
				( 43.49,107.67);
			\end{scope}
			\begin{scope}
			\path[clip] (  0.00,  0.00) rectangle (238.49,115.63);
			\definecolor{drawColor}{gray}{0.30}
			
			\node[text=drawColor,anchor=base east,inner sep=0pt, outer sep=0pt, scale=  0.88] at ( 39.96, 48.03) {1};
			
			\node[text=drawColor,anchor=base east,inner sep=0pt, outer sep=0pt, scale=  0.88] at ( 39.96, 76.10) {100};
			
			\node[text=drawColor,anchor=base east,inner sep=0pt, outer sep=0pt, scale=  0.88] at ( 39.96,104.18) {10000};
			\end{scope}
			\begin{scope}
			\path[clip] (  0.00,  0.00) rectangle (238.49,115.63);
			\definecolor{drawColor}{gray}{0.20}
			
			\path[draw=drawColor,line width= 0.6pt,line join=round] ( 42.16, 51.06) --
				( 44.91, 51.06);
			
			\path[draw=drawColor,line width= 0.6pt,line join=round] ( 42.16, 79.13) --
				( 44.91, 79.13);
			
			\path[draw=drawColor,line width= 0.6pt,line join=round] ( 42.16,107.21) --
				( 44.91,107.21);
			\end{scope}
			\begin{scope}
			\path[clip] (  0.00,  0.00) rectangle (238.49,115.63);
			\definecolor{drawColor}{RGB}{0,0,0}
			
			\path[draw=drawColor,line width= 0.6pt,line join=round] ( 44.91, 30.69) --
				(232.99, 30.69);
			
			\path[draw=drawColor,line width= 0.6pt,line join=round] (230.53, 29.26) --
				(232.99, 30.69) --
				(230.53, 32.11);
			\end{scope}
			\begin{scope}
			\path[clip] (  0.00,  0.00) rectangle (238.49,115.63);
			\definecolor{drawColor}{gray}{0.20}
			
			\path[draw=drawColor,line width= 0.6pt,line join=round] ( 77.88, 27.94) --
				( 77.88, 30.69);
			
			\path[draw=drawColor,line width= 0.6pt,line join=round] (126.74, 27.94) --
				(126.74, 30.69);
			
			\path[draw=drawColor,line width= 0.6pt,line join=round] (175.59, 27.94) --
				(175.59, 30.69);
			
			\path[draw=drawColor,line width= 0.6pt,line join=round] (224.44, 27.94) --
				(224.44, 30.69);
			\end{scope}
			\begin{scope}
			\path[clip] (  0.00,  0.00) rectangle (238.49,115.63);
			\definecolor{drawColor}{gray}{0.30}
			
			\node[text=drawColor,anchor=base,inner sep=0pt, outer sep=0pt, scale=  0.88] at ( 77.88, 19.68) {4};
			
			\node[text=drawColor,anchor=base,inner sep=0pt, outer sep=0pt, scale=  0.88] at (126.74, 19.68) {8};
			
			\node[text=drawColor,anchor=base,inner sep=0pt, outer sep=0pt, scale=  0.88] at (175.59, 19.68) {12};
			
			\node[text=drawColor,anchor=base,inner sep=0pt, outer sep=0pt, scale=  0.88] at (224.44, 19.68) {16};
			\end{scope}
			\begin{scope}
			\path[clip] (  0.00,  0.00) rectangle (238.49,115.63);
			\definecolor{drawColor}{RGB}{0,0,0}
			
			\node[text=drawColor,anchor=base,inner sep=0pt, outer sep=0pt, scale=  1.10] at (138.95,  7.64) {$n$};
			\end{scope}
			\begin{scope}
			\path[clip] (  0.00,  0.00) rectangle (238.49,115.63);
			\definecolor{drawColor}{RGB}{0,0,0}
			
			\node[text=drawColor,rotate= 90.00,anchor=base,inner sep=0pt, outer sep=0pt, scale=  1.10] at ( 13.08, 70.41) {time (ms)};
			\end{scope}
			\begin{scope}
			\path[clip] (  0.00,  0.00) rectangle (238.49,115.63);
			
			\path[] ( 42.61, 63.09) rectangle ( 96.11,117.45);
			\end{scope}
			\begin{scope}
			\path[clip] (  0.00,  0.00) rectangle (238.49,115.63);
			\definecolor{drawColor}{RGB}{247,192,26}
			
			\path[draw=drawColor,line width= 0.6pt,line join=round] ( 49.55,104.72) -- ( 61.12,104.72);
			\end{scope}
			\begin{scope}
			\path[clip] (  0.00,  0.00) rectangle (238.49,115.63);
			\definecolor{fillColor}{RGB}{247,192,26}
			
			\path[fill=fillColor] ( 55.33,104.72) circle (  1.96);
			\end{scope}
			\begin{scope}
			\path[clip] (  0.00,  0.00) rectangle (238.49,115.63);
			\definecolor{drawColor}{RGB}{78,155,133}
			
			\path[draw=drawColor,line width= 0.6pt,dash pattern=on 2pt off 2pt ,line join=round] ( 49.55, 90.27) -- ( 61.12, 90.27);
			\end{scope}
			\begin{scope}
			\path[clip] (  0.00,  0.00) rectangle (238.49,115.63);
			\definecolor{fillColor}{RGB}{78,155,133}
			
			\path[fill=fillColor] ( 55.33, 93.32) --
				( 57.98, 88.74) --
				( 52.69, 88.74) --
				cycle;
			\end{scope}
			\begin{scope}
			\path[clip] (  0.00,  0.00) rectangle (238.49,115.63);
			\definecolor{drawColor}{RGB}{37,122,164}
			
			\path[draw=drawColor,line width= 0.6pt,dash pattern=on 4pt off 2pt ,line join=round] ( 49.55, 75.82) -- ( 61.12, 75.82);
			\end{scope}
			\begin{scope}
			\path[clip] (  0.00,  0.00) rectangle (238.49,115.63);
			\definecolor{fillColor}{RGB}{37,122,164}
			
			\path[fill=fillColor] ( 53.37, 73.85) --
				( 57.30, 73.85) --
				( 57.30, 77.78) --
				( 53.37, 77.78) --
				cycle;
			\end{scope}
			\begin{scope}
			\path[clip] (  0.00,  0.00) rectangle (238.49,115.63);
			\definecolor{drawColor}{RGB}{0,0,0}
			
			\node[text=drawColor,anchor=base west,inner sep=0pt, outer sep=0pt, scale=  0.80] at ( 68.06,101.97) {ACP};
			\end{scope}
			\begin{scope}
			\path[clip] (  0.00,  0.00) rectangle (238.49,115.63);
			\definecolor{drawColor}{RGB}{0,0,0}
			
			\node[text=drawColor,anchor=base west,inner sep=0pt, outer sep=0pt, scale=  0.80] at ( 68.06, 87.52) {DEFT};
			\end{scope}
			\begin{scope}
			\path[clip] (  0.00,  0.00) rectangle (238.49,115.63);
			\definecolor{drawColor}{RGB}{0,0,0}
			
			\node[text=drawColor,anchor=base west,inner sep=0pt, outer sep=0pt, scale=  0.80] at ( 68.06, 73.06) {Naive};
			\end{scope}
		\end{tikzpicture}
		\caption{Run times of the \enquote{naive} approach, the approach used in \ac{cpr}, and \ac{deft} for different numbers of arguments $n$ and a proportion $p = 1.0$ of identical potentials. The left plot shows results for non-exchangeable instances and the right plot presents results for exchangeable instances. Both plots use a logarithmic scale.}
		\label{fig:plot-same}
	\end{figure*}
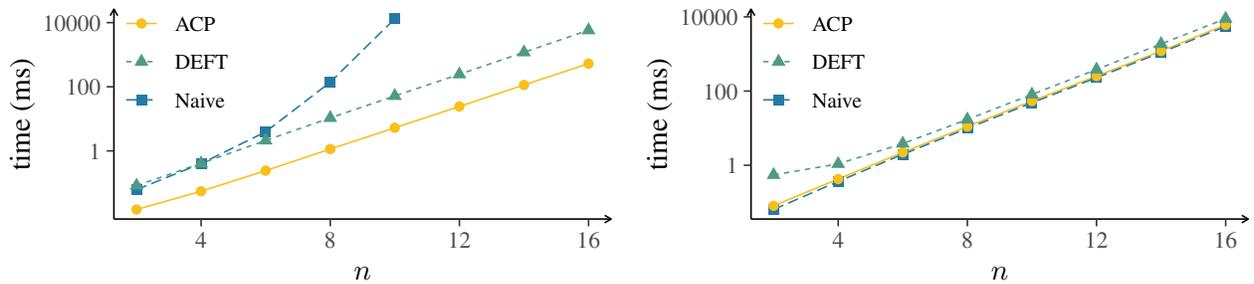

	Finally, \cref{fig:plot-same} presents the run times for instances with a proportion of $p = 1.0$ identical potentials, that is, every input factor maps each assignment of its arguments to the same potential value.
	Even though this scenario is rather unrealistic for practical applications, we include it as an extreme case.
	The left plot again shows similar patterns as the plots for non-exchangeable factors in \cref{fig:plot-asc,fig:plot-mixed}.
	The right plot illustrates that both the naive approach and \ac{cpr} are now able to solve all instances of non-exchangeable factors within the specified timeout.
	All three approaches have a similar run time, which can be explained by the fact that only a single permutation must be considered as the arguments can be rearranged arbitrarily to obtain identical tables (in fact, the tables of the input factors are already identical).
\end{document}